\documentclass[preprint,12pt]{elsarticle}

\usepackage{amssymb}
\usepackage{graphicx, subfigure}
\usepackage{amsfonts, amsmath, amssymb, amsthm, constants, bbm}
\usepackage{MnSymbol}
\usepackage{algorithm, algorithmic}
\usepackage{multirow,booktabs}
\usepackage{tikz}
\usepackage{rotating}

\usepackage{color}
\definecolor{darkred}{RGB}{100,0,0}
\definecolor{darkgreen}{RGB}{0,100,0}
\definecolor{darkblue}{RGB}{0,0,150}

\usepackage{hyperref}
\hypersetup{colorlinks=true, linkcolor=darkred, citecolor=darkgreen, urlcolor=darkblue}
\usepackage{url}

{
\theoremstyle{definition}

}
{
\theoremstyle{remark}
\newtheorem{rem}{Remark}

\newtheorem*{ex*}{Example}

}

\newcommand{\secref}[1]{Section~\ref{sec:#1}}
\newcommand{\figref}[1]{Figure~\ref{fig:#1}}
\newcommand{\algref}[1]{Algorithm~\ref{alg:#1}}

\newcommand{\tabref}[1]{Table~\ref{tab:#1}}

\def\beq{\begin{equation}}
\def\eeq{\end{equation}}
\def\beqn{\begin{eqnarray*}}
\def\eeqn{\end{eqnarray*}}
\def\bitem{\begin{itemize}}
\def\eitem{\end{itemize}}
\def\benum{\begin{enumerate}}
\def\eenum{\end{enumerate}}
\def\bmult{\begin{multline*}}
\def\emult{\end{multline*}}
\def\bcenter{\begin{center}}
\def\ecenter{\end{center}}
\def\bsplit{\begin{split}}
\def\esplit{\end{split}}

\DeclareMathOperator*{\argmax}{arg\, max}
\DeclareMathOperator*{\argmin}{arg\, min}

\def\cC{\mathcal{C}}

\def\cN{\mathcal{N}}

\def\bI{\mathbf{I}}

\newcommand{\bmu}{{\boldsymbol\mu}}

\newcommand\bSigma{{\boldsymbol\Sigma}}

\newcommand{\bdelta}{{\boldsymbol\delta}}

\def\bbD{\mathbb{D}}

\def\bbI{\mathbb{I}}

\def\bbN{\mathbb{N}}

\def\bbR{\mathbb{R}}

\newcommand{\E}{\operatorname{\mathbb{E}}}

\def\symd{\triangle}

\usepackage{dsfont}

\newcommand{\IND}[1]{\bbI\{ #1 \}}

\renewcommand{\paragraph}[1]{\medskip\noindent {\bf #1 \quad}}

\journal{XXXX}
\definecolor{purple}{rgb}{0.4,.1,.9}

\begin{document}

\begin{frontmatter}



\title{A Simple Approach to Sparse Clustering\tnoteref{t1}}
\tnotetext[t1]{Reproducible research. The R code for the numerical experiments is available at\\ 
\url{https://github.com/victorpu/SAS_Hill_Climb} }


\author{Ery Arias-Castro}
    \ead{eariasca@ucsd.edu}
    \ead[url]{http://www.math.ucsd.edu/~eariasca/}
\author{Xiao Pu\corref{cor1}}
    \ead{xipu@ucsd.edu}
    \ead[url]{http://www.math.ucsd.edu/~xipu/}
 \cortext[cor1]{Corresponding author. Fax: (858) 534-5273. Telephone: (858) 705-9107. Address: 9500 Gilman Drive \# 0112, La Jolla, CA  92093-0112, USA}

\address{Department of Mathematics,
University of California, San Diego \\ 9500 Gilman Drive \# 0112, La Jolla, CA  92093-0112, USA}

\begin{abstract}
Consider the problem of sparse clustering, where it is assumed that only a subset of the features are useful for clustering purposes.  In the framework of the COSA method of Friedman and Meulman, subsequently improved in the form of the Sparse K-means method of Witten and Tibshirani, a natural and simpler hill-climbing approach is introduced.  The new method is shown to be competitive with these two methods and others.
\end{abstract}


\end{frontmatter}


\section{Introduction}

Consider a typical setting for clustering $n$ items based on pairwise dissimilarities, with $\delta(i,j)$ denoting the dissimilarity between items $i, j \in [n] := \{1, \dots, n\}$.  For concreteness, we assume that $\delta(i,j) \ge 0$ and $\delta(i,i) = 0$ for all $i, j \in [n]$.
In principle, if we want to delineate $\kappa$ clusters, the goal is (for example) to minimize the average within-cluster dissimilarity.  In detail, a clustering into $\kappa$ groups may be expressed as an assignment function $C: [n] \mapsto [\kappa]$, meaning that $C(i)$ indexes the cluster that observation $i \in [n]$ is assigned to.
Let $\cC^n_\kappa$ denote the class of clusterings of $n$ items into $\kappa$ groups.
For $C \in \cC^n_\kappa$, its average within-cluster dissimilarity is defined as   
\beq
\Delta[C] := \sum_{k \in [\kappa]} \frac{1}{|C^{-1}(k)|} \mathop{\sum\ \sum}_{i,j \in C^{-1}(k)} \delta(i,j).
\label{def:wcd}
\eeq
This dissimilarity coincides with the \emph{within-cluster sum of squares} commonly used in k-means type of clustering algorithms, with $\delta(i,j) = \|x_i-x_j\|^2$. If under the Euclidean setting, we further define cluster centers
\beq
 \mu_k = \frac{1}{n}\sum_{i \in C^{-1}(k)}x_i  \ \ \text{   with } k\in[\kappa],
 \eeq
then the within-cluster dissimilarity can be rewritten as follows,
\beq
\Delta[C] = \sum_{k \in [\kappa]} \frac{1}{|C^{-1}(k)|} \mathop{\sum\ \sum}_{i,j \in C^{-1}(k)}\|x_i-x_j\|^2 = \sum_{k \in [\kappa]} \mathop \sum_{i \in C^{-1}(k)} \|x_i-\mu_k\|^2.
\eeq
Since this paper deals with non-Euclidean settings also, we will use the more general within-cluster dissimilarity defined in \eqref{def:wcd}.  
The resulting optimization problem is the following:
\beq \label{problem0}
\text{Given $(\delta(i,j) : i,j \in [n])$, minimize $\Delta[C]$ over $C \in \cC^n_\kappa$.}
\eeq
This problem is combinatorial and quickly becomes computationally too expensive, even for small datasets.
A number of proposals have been suggested \citep{ESL}, ranging from hierarchical clustering approaches to K-medoids.   
 
Following in the footsteps of \cite{friedman2004clustering}, we consider a situation where we have at our disposal not 1 but $p \ge 2$ measures of pairwise dissimilarities on the same set of items, with $\delta_a(i,j)$ denoting the $a$-th dissimilarity between items $i, j \in [n]$.
Obviously, these measures of dissimilarity could be combined into a single measure of dissimilarity, for example, 
\beq\label{delta-sum}
\delta(i,j) = \sum_a \delta_a(i,j).
\eeq
Our working assumption, however, is that only a few of these measures of dissimilarity are useful for clustering purposes, but we do not know which ones.
This is the setting of sparse clustering, where the number of useful measures is typically small compared to the whole set of available measures.
  
We assume henceforth that all dissimilarity measures are equally important (for example, when we do not have any knowledge a priori on the relative importance of these measures) and that they all satisfy
\beq \label{norm}
\sum_{i,j \in [n]} \delta_a(i,j) = 1, \quad \forall a \in [p],
\eeq
which, in practice, can be achieved via normalization, meaning,
\beq
\delta_a(i,j) \gets \frac{\delta_a(i,j)}{\sum_{i,j} \delta_a(i,j)}.
\eeq
This assumption is important when combining measures in the standard setting \eqref{delta-sum} and in the sparse setting \eqref{delta-S} below.

Suppose for now that we know that at most $s$ measures are useful among the $p$ measures that we are given.  
For $S \subset [p]$, define the $S$-dissimilarity as
\beq\label{delta-S}
\delta_S(i,j) = \sum_{a \in S} \delta_a(i,j),
\eeq
and the corresponding average within-cluster $S$-dissimilarity for the cluster assignment $C$ as
\beq
\Delta_S[C] := \sum_{k \in [\kappa]} \frac{1}{|C^{-1}(k)|} \mathop{\sum\ \sum}_{i,j \in C^{-1}(k)} \delta_S(i,j).
\eeq
If the goal is to delineate $\kappa$ clusters, then a natural objective is the following:
\beq \label{problem}
\begin{array}{c}
\text{Given $(\delta_a(i,j) : a \in [p], i,j \in [n])$,} \\
\text{minimize $\Delta_S[C]$ over $S \subset [p]$ of size $s$ and over $C \in \cC^n_\kappa$.}
\end{array}\eeq
In words, the goal is to find the $s$ measures (which play the role of features in this context) that lead to the smallest optimal average within-cluster dissimilarity.
The problem stated in \eqref{problem} is at least as hard as the problem stated in \eqref{problem0}, and in particular, is computationally intractable even for small item sets.  

\begin{rem}
In many situations, but not all, $p$ measurements of possibly different types are taken from each item $i$, resulting in a vector of measurements $x_i = (x_{ia} : a \in [p])$.  This vector is not necessarily in a Euclidean space, although this is an important example --- see \secref{intro-euclidean} below.  We recover our setting when we have available a dissimilarity measure $\delta_a(i,j)$ between $x_{ia}$ and $x_{ja}$.  This special case justifies our using the terms `feature' and `attribute' when referring to a dissimilarity measure.
\end{rem}



\section{Related work} \label{sec:related}

The literature on sparse clustering is much smaller than that of sparse regression or classification.  Nonetheless, it is substantial and we review some of the main proposals in this section.  We start with the contributions of \cite{friedman2004clustering} and \cite{witten2010framework}, which inspired this work.

\subsection{COSA, Sparse K-means and Regularized K-means}
\cite{friedman2004clustering} propose clustering objects on subsets of attributes (COSA), which (in its simplified form) amounts to the following optimization problem
\begin{gather} 
\text{minimize } \sum_{k \in [\kappa]} \alpha(|C^{-1}(k)|) \sum_{i,j \in C^{-1}(k)} \sum_{a \in [p]} (w_a \delta_a(i,j) + \lambda w_a \log w_a), \label{cosa-full} \\
\text{over any clustering $C$ and any weights $w_1, \dots, w_p \ge 0$ subject to } \sum_{a \in [p]} w_a =1. \label{cosa-feasible}
\end{gather}
Here $\alpha$ is some function and $\lambda \geq 0$ is a tuning parameter. 
When $\alpha(u) = 1/u$, the objective function can be expressed as
\beq\label{cosa}
\sum_{a \in [p]} (w_a \Delta_a[C] + \lambda w_a \log w_a).
\eeq
When $\lambda = 0$, the minimization of \eqref{cosa} over \eqref{cosa-feasible} results in any convex combination of attributes with smallest average within-cluster dissimilarity. If this smallest dissimilarity is attained by only one attribute, then all weights will concentrate on this attribute, with weights $1$ for this attribute and $0$ for the others.
In general, $\lambda > 0$, and the term it multiplies is the negative entropy of the weights $(w_a : a \in [p])$ seen as a distribution on $\{1,\dots,p\}$.
This penalty term encourages the weights to spread out over the attributes.
Minimizing over the weights first leads to 
\begin{gather} 
\text{minimize } \Delta_{\rm cosa}[C] := \min_w \sum_{a \in [p]} (w_a \Delta_a[C] + \lambda w_a \log w_a) \\
\text{over any clustering $C$.}
\end{gather}
where the minimum is over the $w$'s satisfying \eqref{cosa-feasible}.  (Note that the $\lambda$ needs to be tuned.)
The minimization is carried out using an alternating strategy where, starting with an initialization of the weights $w$ (say all equal, $w_a = 1/p$ for all $a \in [p]$), the procedure alternates between optimizing with respect to the clustering assignment $C$ and optimizing with respect to the weights.  (There is a closed-form expression for that derived in that paper.) The procedure stops when achieving a local minimum.
\cite{witten2010framework} observe that an application of COSA rarely results in a sparse set of features, meaning that the weights are typically spread out.
They propose an alternative method, which they call Sparse K-means, which, under \eqref{norm}, amounts to the following optimization problem
\begin{gather}
\text{maximize } \sum_{a \in [p]} w_a \left(\tfrac{1}{n} - \Delta_a[C] \right), \\
\text{over any clustering $C$ and any weights $w_1, \dots, w_p \ge 0$} \\
\text{with } \|w\|_2 \le 1, \ \|w\|_1\le s.
\label{eq:WT's condition}
\end{gather} 
The $\ell_1$ penalty on $w$ results in sparsity for small values of the tuning parameter $s$, which is tuned by the gap statistic of \cite{tibshirani2001estimating}.  
The $\ell_2$ penalty is also important, as without it, the solution would put all the weight on only one the attribute with smallest average within-cluster dissimilarity.
A similar minimization strategy is proposed, which also results in a local optimum.

As can be shown in later sections, Sparse K-means is indeed effective in practice. However, its asymptotic consistency remains unknown.  \cite{sun2012regularized} propose Regularized K-means clustering for high-dimensional data and prove its asymptotic consistency. This method aims at minimizing a regularized \emph{within-cluster sum of squares} with an adaptive group lasso penalty term on the cluster centers:
\begin{gather}
\text{minimize } \frac{1}{n}\sum_{k \in [\kappa]} \mathop \sum_{i \in C^{-1}(k)} \|x_i-\mu_k\|^2 + \sum_{a \in [p]} \lambda_a \sqrt{\mu_{1a}^2 + \cdots + \mu_{\kappa a}^2}~, \\
\text{over  any clustering $C$ and any sets of centers $\mu_1, \mu_2,\cdots,\mu_\kappa$. }
\end{gather}

\subsection{Some methods for the Euclidean setting}
\label{sec:intro-euclidean}
Consider points in space (denoted $x_1, \dots, x_n$ in $\bbR^p$) that we want to cluster.  A typical dissimilarity is the Euclidean metric, denoted by $\delta(i,j) = \|x_i - x_j\|^2$.  Decomposing this into coordinate components, with $x_i = (x_{ia} : a \in [p])$, and letting $\delta_a(i,j) = (x_{ia} - x_{ja})^2$, we have
\beq
\delta(i,j) = \sum_{a \in [p]} \delta_a(i,j).
\eeq
A normalization would lead us to consider a weighted version of these dissimilarities.  But assuming that the data has been normalized to have (Euclidean) norm 1 along each coordinate, \eqref{norm} holds and we are within the framework described above.

This Euclidean setting has drawn most of the attention.
Some papers propose to perform clustering after reducing the dimensionality of the data \citep{ghosh2002mixture, liu2002, tamayo2007metagene}.  However, the preprocessing step of dimensionality reduction is typically independent of the end goal of clustering, making such approaches non-competitive.

A model-based clustering approach is based maximizing the likelihood.  Under the sparsity assumption made here, the likelihood is typically penalized.  Most papers assume a Gaussian mixture model.  Let $f(x; \mu, \bSigma)$ denote the density of the normal distribution with mean $\mu$ and covariance matrix $\bSigma$.  The penalized negative log-likelihood (when the goal is to obtain $\kappa$ clusters) is of the form
\beq\label{likelihood} 
- \sum_{i \in [n]} \log\Big[\sum_{k \in [\kappa]} \pi_k f_k(x_i;\mu_k,\bSigma_k)\Big] + p_\lambda(\Theta),
\eeq
where $\Theta$ gathers all the parameters, meaning, the mixture weights $\pi_1, \dots, \pi_\kappa$, the group means $\mu_1, \dots, \mu_\kappa$, and the group covariance matrices $\bSigma_1, \dots, \bSigma_\kappa$.
For instance, assuming that the data has been standardized so that each feature has sample mean $0$ and variance $1$, \cite{pan2007penalized} use
\beq\label{pan-shen}
p_\lambda(\Theta)=\lambda \sum_{k \in [\kappa]} \|\mu_k\|_1.
\eeq
This may be seen as a convex relaxation of
\beq
p_\lambda(\Theta)=\lambda \sum_{a \in [p]} \sum_{k \in [\kappa]} \IND{\mu_{ka} \ne 0} = \lambda \sum_{k \in [\kappa]} \|\mu_k\|_0.
\eeq
Typically, this optimization will result in some coordinates set to zero and thus deemed not useful for clustering purposes.
In another variant, \cite{wang2008variable} use
\beq
p_\lambda(\Theta)=\lambda \sum_{a \in [p]} \max_{k \in [\kappa]} |\mu_{ka}|.
\eeq
To shrink the difference between every pair of cluster centers for each variable $a$, \cite{guo2010pairwise} use the \emph{pairwise fusion penalty
\beq
p_\lambda(\Theta) = \lambda \sum_{a \in [p]}\sum_{1\leq k\leq k'\leq\kappa}|\mu_{ka}-\mu_{k'a}|.
\eeq}
Taking into account the covariance matrices, and assuming they are diagonal, \cite{xie2008penalized} use
\beq
p_\lambda(\Theta)=\lambda_1\sum_{k \in [\kappa]}\sum_{a \in [p]}|\mu_{ka}| +\lambda_2\sum_{k \in [\kappa]}\sum_{a \in [p]}|\sigma_{ka}^2-1|.
\eeq
The assumption that the covariance matrices are diagonal is common in high-dimensional settings and was demonstrated to be reasonable in the context of clustering  \citep{fraley2006mclust}.  
Note that none of these proposals make the optimization problem \eqref{likelihood} convex or otherwise tractable.  The methods are implemented via an EM-type approach.

Another line of research on sparse clustering is based on coordinate-wise testing for mixing.  This constitutes the feature selection step.  The clustering step typically amounts to applying a clustering algorithm to the resulting feature space.
For example, \cite{jin2014important} use a Kolmogorov-Smirnov test against the normal distribution, while \cite{jin2015phase} use a (chi-squared) variance test.
The latter is also done in \citep{azizyan2013} and in \citep{verzelen2014detection}.
This last paper also studies the case where the covariance matrix is unknown and proposes an approach via moments.  
In a nonparametric setting, \cite{chan2010using} use coordinate-wise mode testing.


\section{Our method: Sparse Alternate Sum (SAS) Clustering}
  
Hill-climbing methods are iterative in nature, making `local', that is, `small' changes at each iteration.  They have been studied in the context of graph partitioning, e.g., by \cite{kernighan1970efficient} and \cite{carson2001hill}, among others.  
In the context of sparse clustering, we find the K-medoids variant of \cite{aggarwal1999fast}, which includes a hill-climbing step.

Many of the methods cited in \secref{related} use alternate optimization in some form (e.g., EM), which can be interpreted as hill-climbing.  
Our method is instead directly formulated as a hill-climbing approach, making it simpler and, arguably, more principled than COSA or Sparse K-means.   

\subsection{Our approach: SAS Clustering}
\label{sec:sas-clustering}
Let $\hat C$ be an algorithm for clustering based on dissimilarities.  Formally, $\hat C : \bbD \times \bbN \mapsto \cC$, where $\bbD$ is a class of dissimilarity matrices and $\cC := \bigcup_n \bigcup_\kappa \cC^n_\kappa$, and for $(\delta, \kappa) \in \bbD \times \bbN$ with $\delta$ of dimension $n$, $\hat C(\delta, \kappa) \in \cC^n_\kappa$.  
Note that $\hat C$ could be a hill-climbing method for graph partitioning, or K-medoids (or K-means if we are provided with points in a vector space rather than dissimilarities), or a spectral method, namely, any clustering algorithm that applies to dissimilarities. 
(In this paper, we will use K-means for numerical data and K-medoids for categorical data using hamming distances as dissimilarities.)
For $S \subset [p]$, define
\beq\label{bdelta}
\bdelta_S = (\delta_a(i,j) : a \in S; i,j \in [n]) 
\quad \text{and} \quad 
\bdelta = \bdelta_{[p]}.
\eeq
Our procedure is described in \algref{sasc}.

\begin{algorithm}[!th]
\caption{\quad Sparse Alternate Similarity (SAS) Clustering}
\label{alg:sasc}
\begin{algorithmic}
\STATE {\bf Input:} dissimilarities $(\delta_a(i,j) : a \in [p], i,j \in [n])$, number of clusters $\kappa$, number of features $s$
\STATE {\bf Output:} feature set $S$, group assignment function $C$
\STATE {\bf Initialize:} For each $a \in [p]$, compute $C_a \gets \hat C(\bdelta_a, \kappa)$ and then $\Delta_a[C_a]$.  Let $S \subset [p]$ index the smallest $s$ among these.
\STATE {\bf Alternate} between the following steps until `convergence':
\STATE {\bf 1:} Keeping $S$ fixed, compute $C \gets \hat C(\bdelta_S, \kappa)$.
\STATE {\bf 2:} Keeping $C$ fixed, compute $S \gets \argmin_{|S| = s} \Delta_S[C]$.
\end{algorithmic}
\end{algorithm}

The use of algorithm $\hat C$ in Step 1 is an attempt to minimize $C \mapsto \Delta_S[C]$ over $C \in \cC^n_\kappa$.
The minimization in Step 2 is over $S \subset [p]$ of size $s$ and it is trivial.  Indeed, the minimizing $S$ is simply made of the $s$ indices $a \in [p]$ corresponding to the smallest $\Delta_a[C]$. 
For the choice of parameters $\kappa$ and $s$, any standard method for tuning parameters of a clustering algorithm applies, for example, by optimization of  the gap statistic of \cite{tibshirani2001estimating}.
We note that the initialization phase, by itself, is a pure coordinate-wise approach that has analogs in the Euclidean setting as mentioned \secref{intro-euclidean}.
The hill-climbing process is the iteration phase. 

\begin{rem}
We tried another initialization in \algref{sasc} consisting of drawing a feature set $S$ at random.  We found that the algorithm behaved similarly.  (Results not reported here.)
\end{rem}

Compared with COSA and Sparse K-means, and other methods based on penalties, we note that the choice of features in our SAS algorithm is much simpler, using a hill-climbing approach instead.

\subsection{Number of iterations needed}
\label{sec:iterations}
\begin{figure}[h]
\centering\footnotesize
\subfigure{
\begin{tikzpicture}
\node at (0,2.2) {$\mu = 0.6$};
\node at (-3.3,0) [ rotate=90] {\textcolor{green}{Rand index}};
\node at (0,0) {\includegraphics  [scale=0.25]{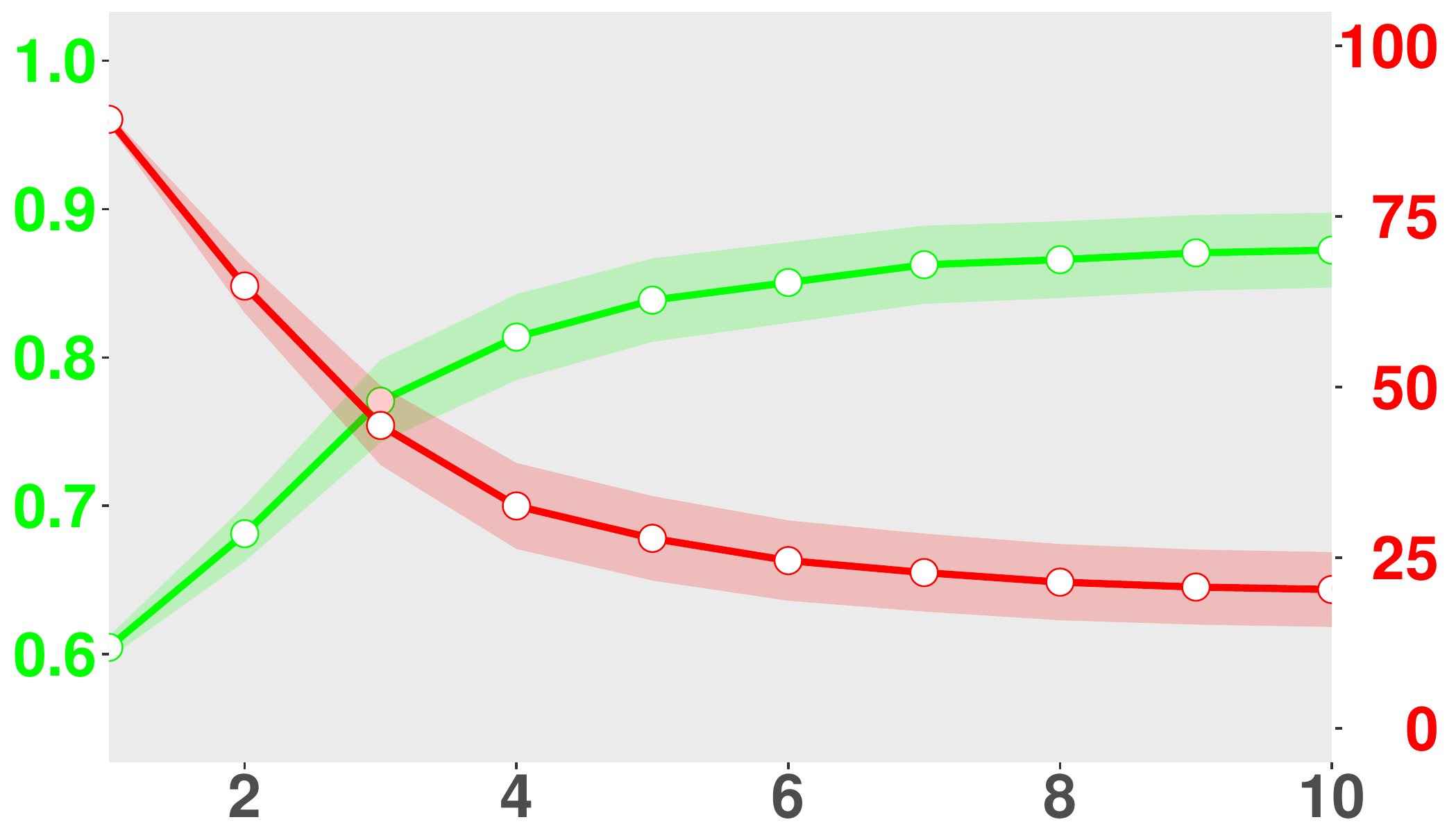}};
\end{tikzpicture}}
~
\subfigure{
\begin{tikzpicture}
\node at (0,2.2) {$\mu = 0.7$};
\node at (3.3,0) [rotate=90] {\textcolor{red}{$|S_* \symd \hat S|$}};
\node at (0,0) {\includegraphics  [scale=0.25]{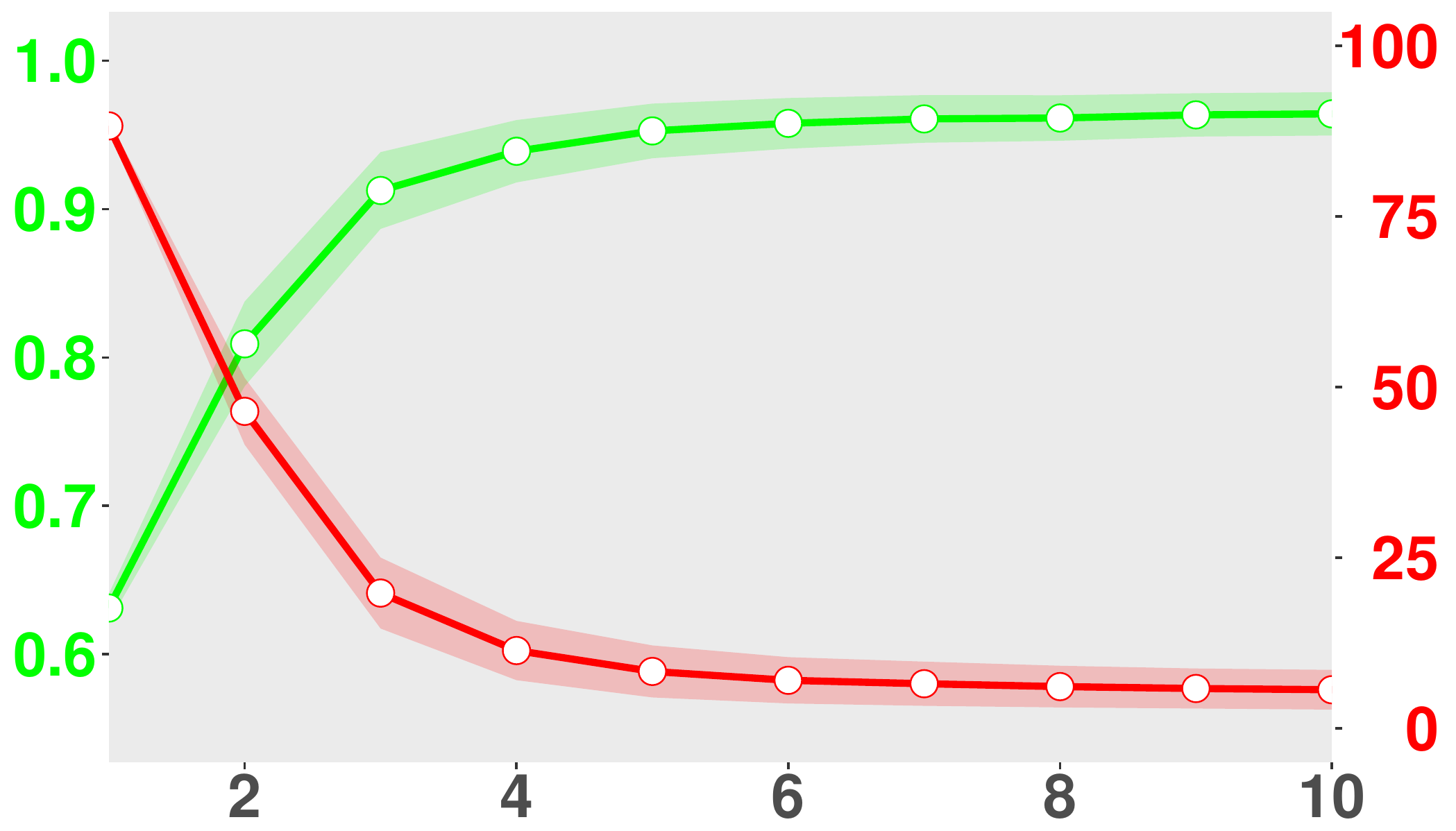}};
\end{tikzpicture}}
~	
\subfigure{
\begin{tikzpicture}
\node at (0,2.2){$\mu = 0.8$};
\node at (-3.3,0) [rotate=90] {\textcolor{green}{Rand index}};
\node at (0,0){\includegraphics  [scale=0.25]{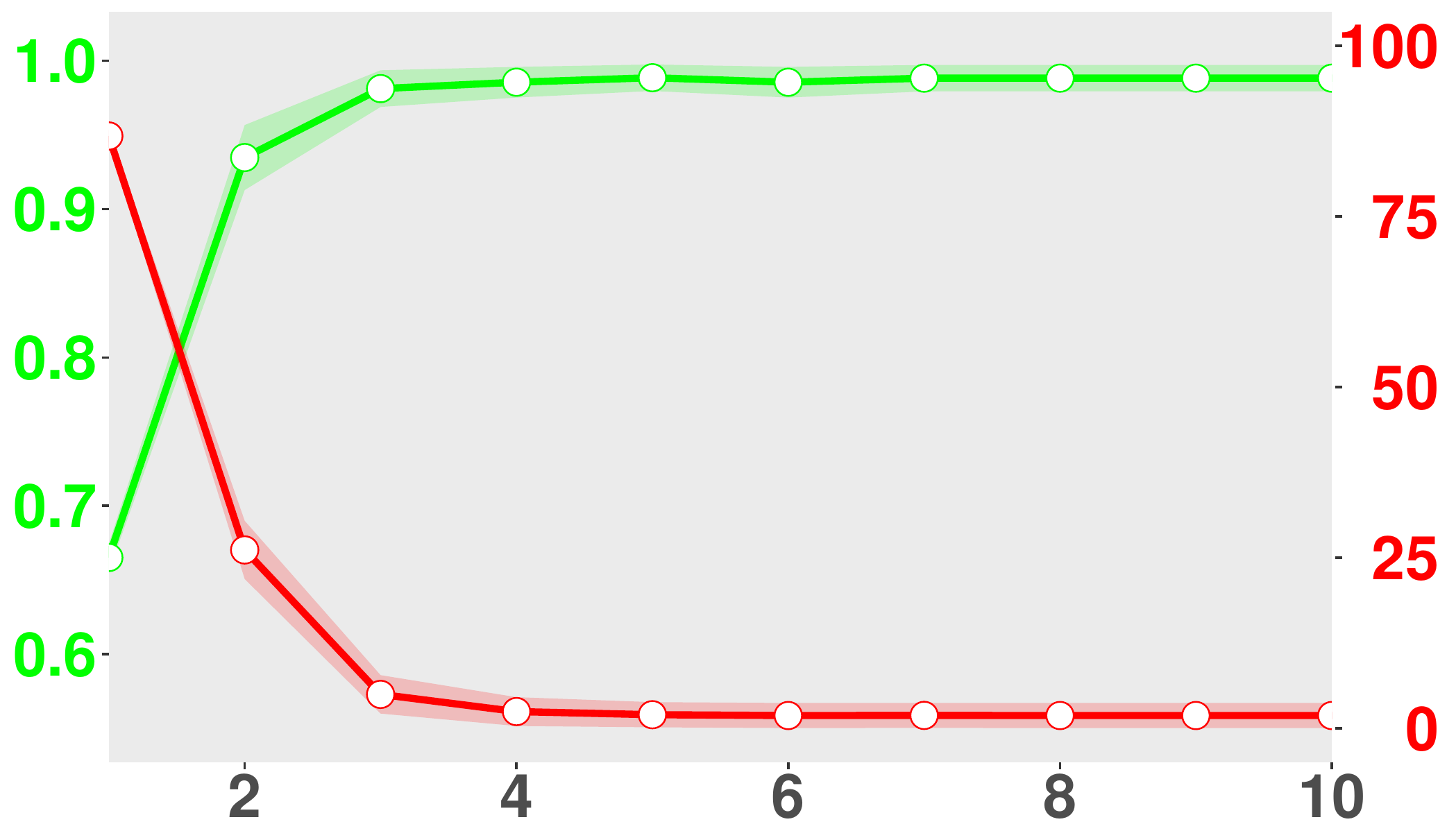}};
\end{tikzpicture}}
~
\subfigure{
\begin{tikzpicture}
\node at (0,2.2){$\mu = 0.9$};
\node at (3.3,0) [rotate=90] {\textcolor{red}{$|S_* \symd \hat S|$}};
\node at (0,0){\includegraphics  [scale=0.25]{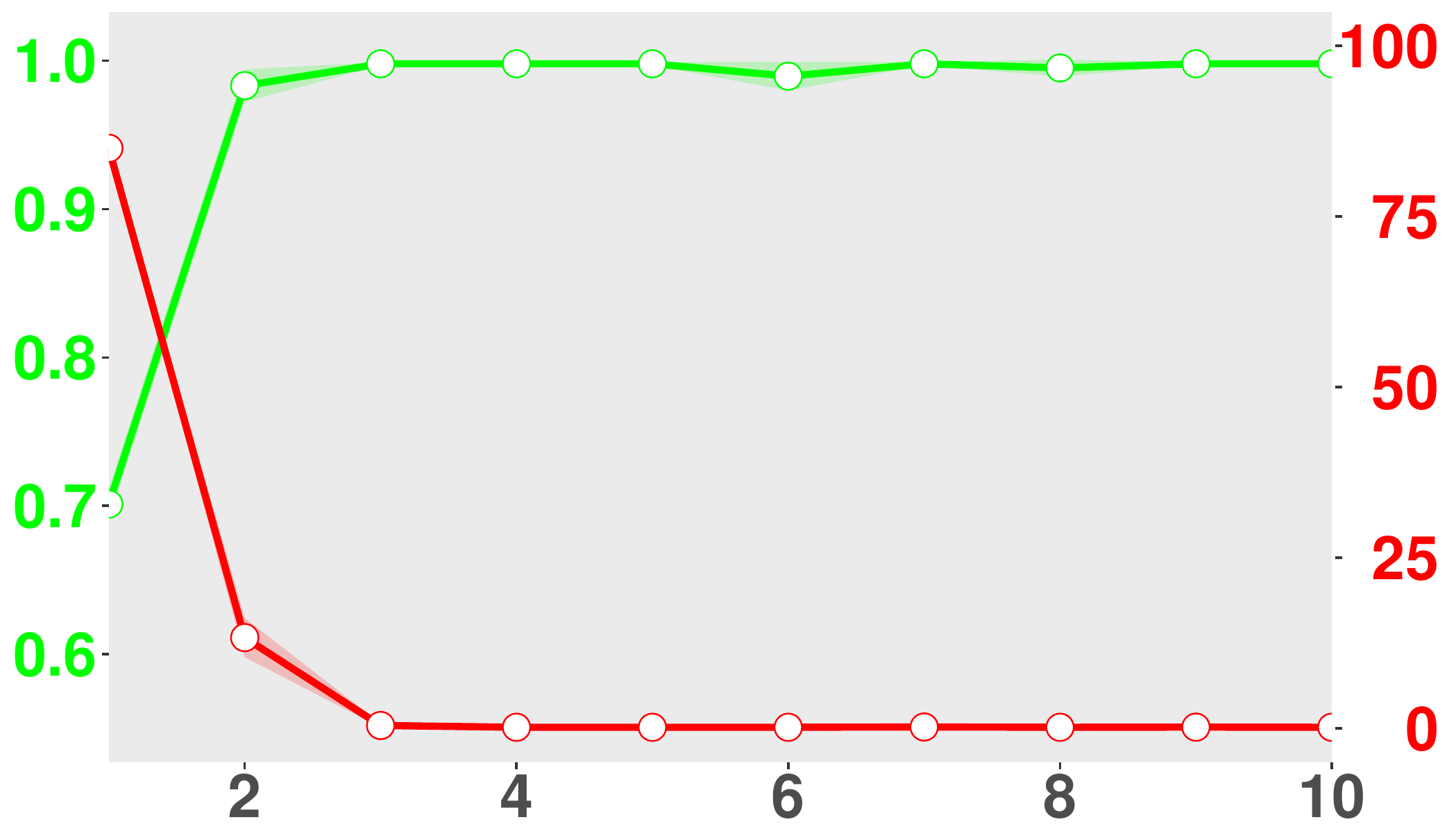}};
\end{tikzpicture}}

\begin{tikzpicture}
\node at (-3.2,0){Iteration number};
\node at (3.5,0){Iteration number};
\end{tikzpicture}
\caption{Means (and 95\% confidence intervals of the means) of Rand indexes and symmetric differences.  
}
\label{fig:convergence}
\end{figure}
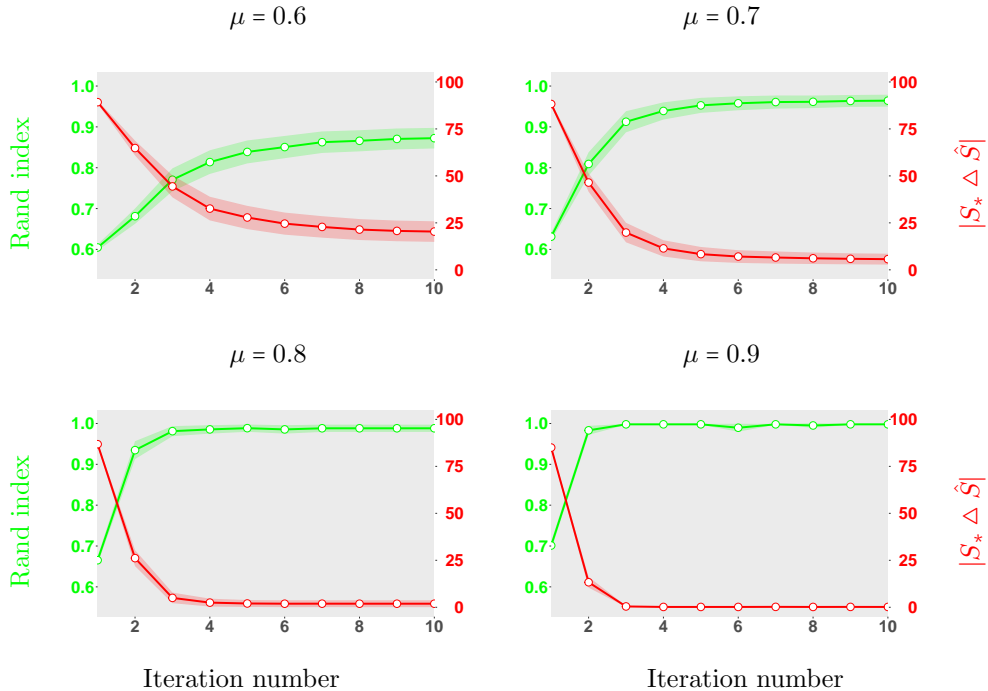

A first question of interest is whether the iterations improve the purely coordinate-wise method, defined as the method that results from stopping after one pass through Steps 1-2 in \algref{sasc} (no iteration).
Although this is bound to vary with each situation, we examine an instance where the data come from the mixture of three Gaussians with sparse means.
In detail, the setting comprises 3 clusters with 30 observations each and respective distributions $\cN(\bmu,\bI)$, $\cN({\bf 0},\bI)$ and $\cN(-\bmu,\bI)$, with $\bmu = (\mu, \dots, \mu, 0, \dots, 0)$ having 50 $\mu$'s and 450 zeros. We assume that $\kappa = 3$ and $s = 50$ are both given, and we run the SAS algorithm and record the Rand indexes \citep{rand1971objective} and symmetric differences $|S_* \symd \hat S|$ as the end of each iteration of Steps 1-2. 
The setting is repeated 400 times.
The means and confidence intervals under different regimes ($\mu = 0.6$, $\mu = 0.7$, $\mu = 0.8$, $\mu = 0.9$) are shown in \figref{convergence}. 
At least in this setting, the algorithm converges in a few iterations and, importantly, these few iterations bring significant improvements, particularly over the purely coordinate-wise algorithm. 

\subsection{Selection of the sparsity parameter}
We consider the problem of selecting $\kappa$, the number of clusters, as outside of the scope of this work, as it is intrinsic to the problem of clustering and has been discussed extensively in the literature --- see \citep{tibshirani2001estimating,kou2014estimating} and references therein.  
Thus we assume that $\kappa$ is given.
Besides $\kappa$, our algorithm has one tuning parameter, the sparsity parameter $s$, which is the number of useful features for clustering, meaning, the cardinality of set $S$ in \eqref{problem}. 

Inspired by the gap statistic of \cite{tibshirani2001estimating}, which was designed for selecting the number of clusters $\kappa$ in standard K-means clustering, we propose a permutation approach for selecting $s$. 
Let $\Delta_s^{\rm obs}$ denote the average within-cluster dissimilarity of the clustering computed by the algorithm on the original data with input number of features $s$.
Let $\Delta_s^{\rm perm}$ denote the same quantity but obtained from a random permutation of the data --- a new sample is generated by independently permuting the observations within each feature.  
The gap statistic (for $s$) is then defined as
\beq
\text{gap}(s) = \log\Delta_s^{\rm obs} - \E(\log\Delta_s^{\rm perm}).
\eeq
In practice, the expectation is estimated by Monte Carlo, generating $B$ random permuted datasets.
A large gap statistic indicates a large discrepancy between the observed amount of clustering and that expected of a null model (here a permutation of the data) with no salient clusters. 

The optimization of the gap statistics over $s \in [p]$ is a discrete optimization problem.  An exhaustive search for $s$ would involve computing $p$ gap statistics, each requiring $B$ runs of the SAS algorithm.  
This is feasible when $p$ and $B$ are not too large.\footnote{In our experiments, we choose $B = 25$ as in the code that comes with \citep{witten2010framework}.}   
See \algref{sasgs}, which allows for coarsening the grid.

\begin{algorithm}[!th]
\caption{\quad SAS Clustering with Grid Search}
\label{alg:sasgs}
\begin{algorithmic}
\STATE {\bf Input:} Dissimilarities $(\delta_a(i,j) : a \in [p], i,j \in [n])$, number of clusters $\kappa$, step size $h$, number of Monte Carlo permutations $B$
\STATE {\bf Output:} Number of useful features $\hat s$, feature set $S$, group assignment  $C$
\FOR {$s = 1 \text{ to } p$ with step size $h$}
\STATE {Run {\bf Algorithm 1} to get the feature set $S_s$ and group assignment $C_s$}
\STATE {Run {\bf Algorithm 1} on $B$ permuted datasets to get the gap statistic $G_s$}
\ENDFOR
\RETURN Let $\hat s = \argmax_s G_s$ and return $S_{\hat s}$ and $C_{\hat s}$ \\
\end{algorithmic}
\end{algorithm}
 
\begin{figure}[h]
\centering\footnotesize
\begin{tikzpicture}
\node at (0,0) {\includegraphics[scale=0.35]{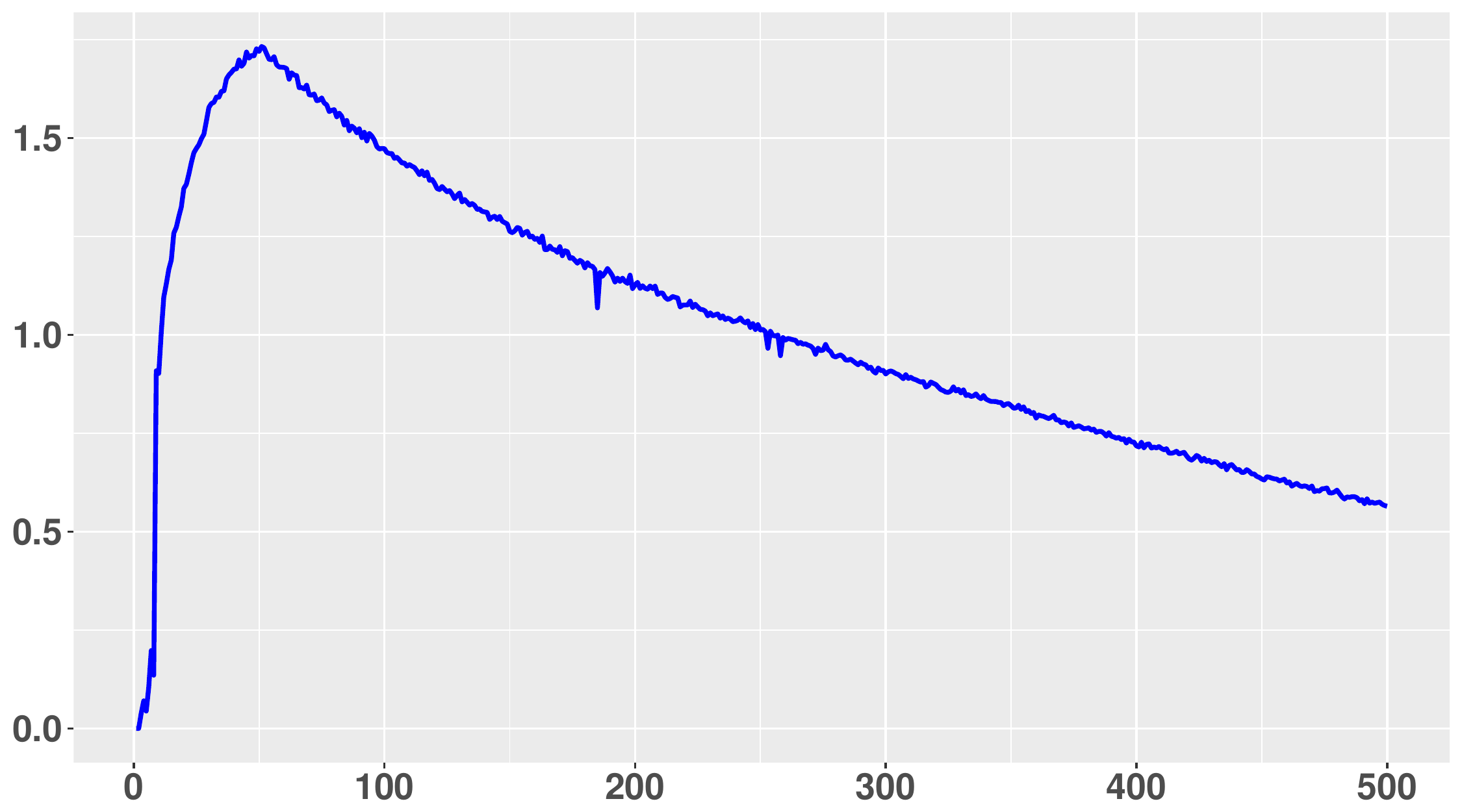}};
\node at (0,-2.6) [scale=1] {sparsity parameter $s$};
\node at (-4.3,0) [scale=1, rotate=90] {gap statistic};
\end{tikzpicture}
\caption{A plot of the gap statistic for each $s \in [p]$ for a Gaussian mixture with 3 components (30 observations in each cluster) in dimension $p = 500$.}
\label{fig:gaps}
\end{figure}

To illustrate the effectiveness of choosing $s$ using the gap statistic, we computed the gap statistic for all $s \in [p]$ in the same setting as that of \secref{iterations} with $\mu = 1$.  
The result of the experiment is reported in \figref{gaps}.
Note that, in this relatively high SNR setting, the gap statistic achieves its maximum at the correct number of features. 


In this experiment, at least, the gap statistic seems unimodal (as a function of $s$).  If it were the case, we could use a golden section search, which would be much faster than an exhaustive grid search.

\section{Numerical experiments}
\label{sec:comparison}

We performed a number of numerical experiments, both on simulated data and on real (microarray) data to compare our method with other proposals. Throughout this section, we standardize the data coordinate-wisely, we assume that the number of clusters is given, and we use the gap statistic of \cite{tibshirani2001estimating} to choose the tuning parameter $s$ in our algorithm.

\subsection{A comparison of SAS Clustering with Sparse K-means and IF-PCA-HCT} \label{sec:sim_identity}

We compare our \algref{sasc} with IF-PCA-HCT \citep{jin2014important} and Sparse K-means \citep{witten2010framework} in the setting of \secref{iterations}.  We note that IF-PCA-HCT was specifically designed for that model and that Sparse K-means was shown to numerically outperform a number of other approaches, including standard K-means, COSA \citep{friedman2004clustering}, model-based clustering \citep{raftery2006variable}, the penalized log-likelihood approach of \citep{pan2007penalized} and the classical PCA approach.  
We use the gap statistic to tune the parameters of SAS Clustering and Sparse K-means.  (SAS\_gs uses a grid search while SAS\_gss uses a golden section search.) 
IF-PCA-HCT is tuning-free --- it employs the higher criticism to automatically choose the number of features.

\begin{figure}[h]
\centering\footnotesize
\begin{tikzpicture}
\node at (0,0) {\includegraphics[scale=0.35]{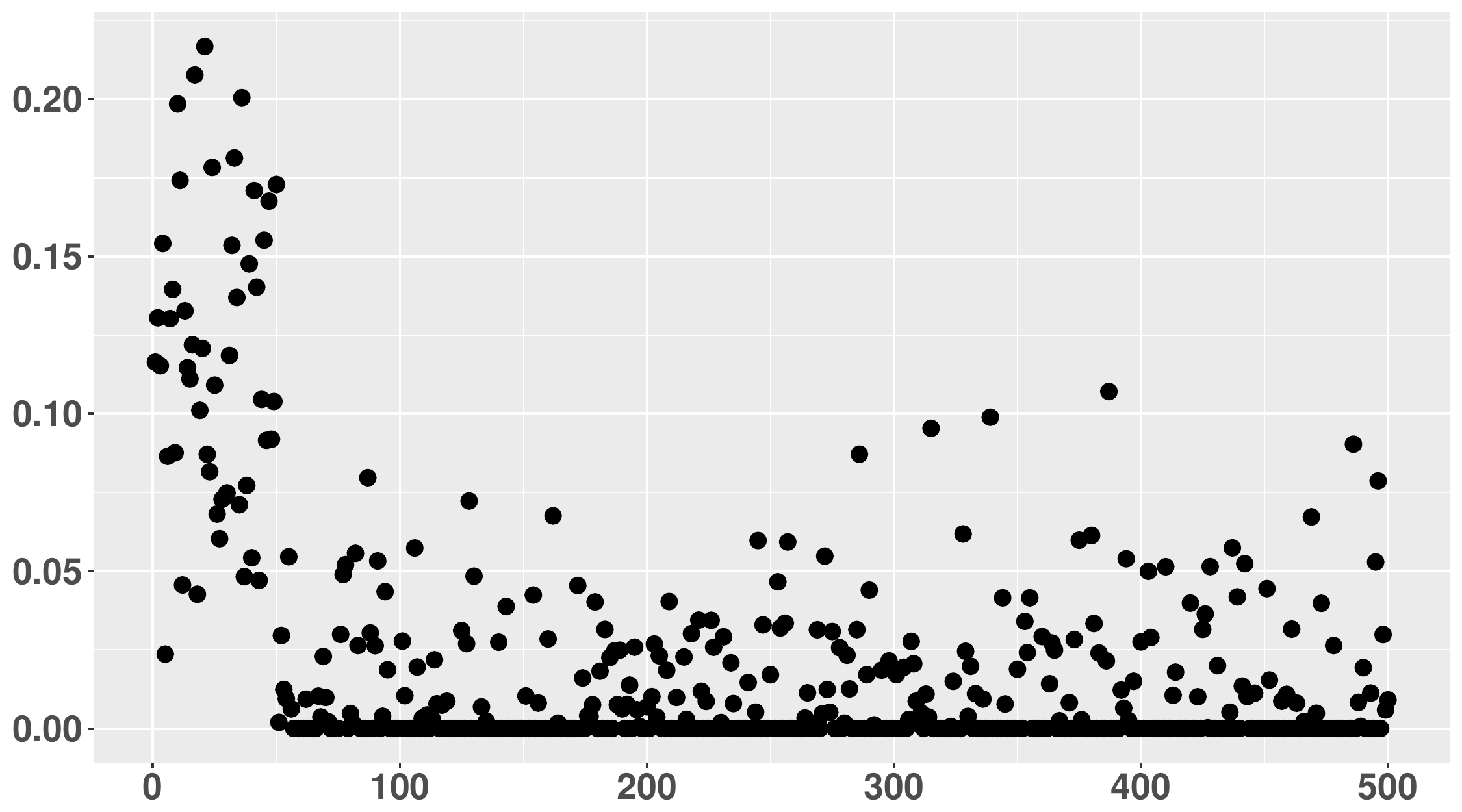}};
\node at (0,-2.6) [scale=1] {Feature Index};
\node at (-4.3,0) [scale=1, rotate=90] {Weights};
\end{tikzpicture}
\caption{A typical example of the weights that Sparse K-means returns.}
\label{fig:weights_a}
\end{figure}

In \tabref{1a}, we report the performance for these three methods in terms of Rand index \citep{rand1971objective} for various combinations of $\mu$ and $p$. Each situation was replicated 50 times. 
As can be seen from the table, SAS Clustering outperforms IF-PCA-HCT, and performs at least as well as Sparse K-means and sometimes much better (for example when $p = 500$ and $\mu = 0.7$). We examine a dataset from this situation in depth, and plot the weights resulted from Sparse K-means on this dataset, see \figref{weights_a}.
As seen in this figure, and also as mentioned in \citep{witten2010framework}, Sparse K-means generally results in more features with non-zero weights than the truth. These extraneous features, even with small weights, may negatively impact the clustering result. In this specific example, the Rand index from Sparse K-means is 0.763 while our approach gives a Rand index of 0.956.  
Let $S_* \subset [p]$ denote the true feature set and $\hat S$ the feature set that our method return.  In this example, $|S_* \symd \hat S| = 12$.  

While both SAS Clustering and Sparse K-means use the gap statistic to tune the parameters, IF-PCA-HCT tunes itself analytically without resorting to permutation or resampling, and (not surprisingly) has the smallest computational time among these three methods. However, as can be seen from \tabref{1a}, the clustering results given by IF-PCA-HCT are far worse than those resulted from the other two methods. In \tabref{1b}, we report the performance of SAS Clustering and Sparse K-means in terms of the running time, under the same setting as that in \tabref{1a} but with tuning parameters for both of the methods given (so that the comparisons are fair). As can be seen in \tabref{1b}, SAS Clustering shows a clear advantage over Sparse K-means in terms of the running time, and as $p$ increases, the advantage becomes more obvious.
(Note that both SAS and Sparse K-means are implemented in R code and, in particular, the code is not optimized.)

\begin{table}[h]
\renewcommand\thetable{1a}
\centering\footnotesize
\begin{tabular}{lccccc}
\hline
$\mu$ & methods & p = 100 & p = 200& p = 500& p = 1000\\
\hline
\multirow{4}{*}{0.6} 
& SAS\_gs &0.907 (0.048)   & 0.875 (0.066) & 0.827 (0.076) &0.674 (0.096)  \\
& SAS\_gss &0.900 (0.054) & 0.860 (0.066)& 0.781 (0.008) &0.701(0.050) \\
& Sparse K &0.886 (0.068) &0.807 (0.064)&0.744 (0.046) &0.704 (0.043)\\
&IF-PCA&0.664(0.042) &0.645(0.051) & 0.605 (0.045)&0.593(0.038) \\
\hline
\multirow{4}{*}{0.7}
&SAS\_gs&0.953 (0.030)   &0.965 (0.028)& 0.960 (0.032) &0.855 (0.102)\\
&SAS\_gss &0.953 (0.031)&0.961 (0.031) &0.921 (0.088) &0.789 (0.104) \\
&Sparse K &0.942 (0.045)&0.915 (0.071) &0.802 (0.087) &0.790 (0.087) \\
&IF-PCA &0.681(0.036) &0.653(0.044)&0.629(0.057)&0.614(0.055)\\
\hline
\multirow{4}{*}{0.8}
&SAS\_gs &0.986 (0.020)  &0.985 (0.022)&0.987 (0.016) &0.966 (0.052)  \\
&SAS\_gss &0.984 (0.020)&0.983 (0.019) & 0.987 (0.0178)&0.892 (0.122) \\
&Sparse K &0.985 (0.020)&0.975 (0.029) &0.961 (0.07) &0.948 (0.074) \\
&IF-PCA &0.691(0.043) &0.675(0.056) &0.639(0.068) &0.623(0.059)\\
\hline
\multirow{4}{*}{0.9}
&SAS\_gs& 0.997 (0.008) &0.997 (0.008)&0.997 (0.007) &0.995 (0.010) \\
&SAS\_gss &0.996 (0.010)& 0.996 (0.009)&0.997 (0.009) &0.969 (0.076) \\
&Sparse K &0.996 (0.010)&0.992 (0.013) &0.992(0.016) &0.993 (0.013)\\
&IF-PCA &0.700(0.031) &0.682(0.051) & 0.654(0.057)&0.627(0.065)\\
\hline
\multirow{4}{*}{1.0}
&SAS\_gs &0.999 (0.005)&1.000 (0.003)   &1.000 (0.003)& 0.999 (0.004)   \\
&SAS\_gss &0.998 (0.007)&1.000 (0.003)   &1.000 (0.004) & 0.998 (0.006)  \\
&Sparse K &0.998 (0.007)&0.999 (0.005)   &0.996 (0.010)&0.996 (0.009) \\
&IF-PCA &0.717(0.034) &0.710(0.039) &0.659(0.063)&0.639(0.060)\\
\hline
\end{tabular}
\caption{Comparison results for the simulations in \secref{sim_identity}. The reported values are the mean (and sample standard deviation) of the Rand indexes over 50 simulations. 
}
\label{tab:1a}
\end{table}

\begin{table}[h]
\renewcommand\thetable{1b}
\centering\footnotesize
\begin{tabular}{lccccc}
\hline
$\delta_\mu$ & methods & p = 100 & p = 200& p = 500& p = 1000\\
\hline
\multirow{2}{*}{0.6} 
& SAS &0.086 (0.031)   & 0.130 (0.044) & 0.217 (0.088) &0.271 (0.118)  \\
& Sparse K &0.113 (0.034) &0.220 (0.053)&0.445 (0.101) &0.850 (0.156)\\
\hline
\multirow{2}{*}{0.7} 
& SAS &0.077 (0.021)   & 0.104 (0.027) & 0.207 (0.085) &0.316 (0.147)  \\
& Sparse K &0.107 (0.028) &0.235 (0.057)&0.471 (0.123) &0.945 (0.194)\\
\hline
\multirow{2}{*}{0.8}
&SAS&0.056 (0.019)   &0.088 (0.022)& 0.182 (0.062) &0.313 (0.118)\\
&Sparse K &0.091 (0.029)&0.213 (0.051) &0.574 (0.134) &0.984 (0.262) \\
\hline
\multirow{2}{*}{0.9}
&SAS&0.055 (0.017)   &0.080 (0.024)& 0.136 (0.048) &0.289 (0.131)\\
&Sparse K &0.094 (0.023)&0.196 (0.052) &0.482 (0.101) &0.982 (0.261) \\
\hline
\multirow{2}{*}{1.0}
&SAS &0.051(0.012)&0.089 (0.021)  &0.146 (0.045)&0.272 (0.095)   \\
&Sparse K&0.095(0.019) &0.186 (0.044)&0.554 (0.107) &1.225 (0.270) \\
\hline
\end{tabular}

\caption{Comparison of running time of SAS Clustering (with the number of features $s$ given) and Sparse K-means (with known tuning parameter $s$ in \eqref{eq:WT's condition}) in the setting of \secref{sim_identity}.  Reported is the averaged running time (in seconds)  over 100 repeats, with sample standard deviation in parentheses.}
\label{tab:1b}
\end{table}
\addtocounter{table}{-1}

\subsection{A more difficult situation (same covariance)} 
\label{sec:sim_same_sigma}

In \secref{sim_identity}, the three groups had identity covariance matrix.  In this section, we continue comparing our approach with Sparse K-means and IF-PCA-HCT under a more difficult situation, where each of the 3 clusters have 30 points sampled from different $p$-variate normal distributions ($p =100,200,500,1000$), with different mean vectors
\[
\bmu_1 =[1.02, 1.04, ... , 2, \underbrace{0, ... ,0}_{p-50\text{ zeros}}], \quad\]
\[ \bmu_2 =[1.02+\delta_\mu, 1.04+\delta_\mu, ... , 2+\delta_\mu, \underbrace{0, ... ,0}_{p-50\text{ zeros}}], \quad\]
\[
\bmu_3 =[1.02+2\delta_\mu, 1.04+2\delta_\mu, ... , 2+2\delta_\mu, \underbrace{0, ... ,0}_{p-50\text{ zeros}}],
\]
and same diagonal covariance matrix $\bSigma$ across groups, a random matrix with eigenvalues in $[1,5]$.
We used 50 repeats and varied $\delta_\mu$ from $0.6$ to $1.0$.  The results are reported in Table \ref{tab:sim_same_sigma_a}. 
We see there that, in this setting, our method is clearly superior to Sparse K-means and IF-PCA-HCT. We also report the symmetric difference $|S_* \symd \hat S|$ between the estimated feature set $\hat S$ and the true feature set $S_*$, as can be seen in Table \ref{tab:sim_same_sigma_b}.  Our algorithm is clearly more accurate in terms of feature selection.

\begin{table}[h]
\renewcommand\thetable{2a}
\centering\footnotesize
\begin{tabular}{lccccc}
\hline
$\delta_\mu$ & methods & p = 100 & p = 200& p = 500& p = 1000\\
\hline
\multirow{4}{*}{0.6} 
& SAS\_gs &0.718 (0.037)   & 0.702 (0.037) & 0.611 (0.044) &0.574 (0.027)  \\
& SAS\_gss &0.714 (0.028) & 0.692 (0.038)& 0.635 (0.044) &0.595(0.026) \\
& Sparse K &0.590 (0.030) &0.594 (0.034)&0.595 (0.034) &0.571 (0.023)\\
&IF-PCA&0.619(0.037) &0.590(0.037) & 0.572 (0.024)&0.564(0.020) \\
\hline
\multirow{4}{*}{0.8}
&SAS\_gs&0.852 (0.047)   &0.844 (0.052)& 0.797 (0.066) &0.670 (0.082)\\
&SAS\_gss &0.848 (0.050)&0.819 (0.070) &0.752 (0.060) &0.686 (0.043) \\
&Sparse K &0.662 (0.057)&0.646 (0.063) &0.657 (0.062) &0.639 (0.054) \\
&IF-PCA &0.646(0.040) &0.634(0.047)&0.603(0.046)&0.575(0.040)\\
\hline
\multirow{4}{*}{1.0}
&SAS\_gs &0.940 (0.035)  &0.947 (0.033)&0.941 (0.037) &0.919 (0.065)  \\
&SAS\_gss &0.935 (0.037)&0.941 (0.038) & 0.922 (0.059)&0.799 (0.099) \\
&Sparse K &0.798 (0.085)&0.814 (0.078) &0.742 (0.080) &0.708 (0.070) \\
&IF-PCA &0.677(0.041) &0.644(0.056) &0.618(0.052) &0.604(0.047)\\
\hline
\end{tabular}

\caption{Comparison of SAS Clustering with Sparse K-means  and IF-PCA in the setting of \secref{sim_same_sigma}.  Reported is the averaged Rand index over 50 repeats, with the standard deviation in parentheses.}
\label{tab:sim_same_sigma_a}
\end{table}

\begin{table}[h]
\renewcommand\thetable{2b}
\centering\footnotesize
\begin{tabular}{lccccc}
\hline
$\delta_\mu$ & methods & p = 100 & p = 200& p = 500& p = 1000\\
\hline
\multirow{4}{*}{0.6} 
& SAS\_gs &26.3(4.7)   & 37.7 (7.9) & 86.2 (20.5) &121.0(28.3)  \\
& SAS\_gss &27.9(5.3) & 44.1 (12.1)& 100.5 (43.3) &143.1(57.0) \\
& Sparse K &43.8 (7.3) &86.8(35.3)&163.9(105.9) &170.6 (124.6)\\
&IF-PCA&49.4(3.8) &72.3(15.8) & 129.7 (61.4)&185.8(126.3) \\
\hline
\multirow{4}{*}{0.8}
&SAS\_gs&17.4(3.9)   &19.0(3.9)& 31.7 (17.2) &94.2 (48.3)\\
&SAS\_gss &17.7 (4.6)&21.9 (6.5) &57.9 (43.7) &132.9 (85.0) \\
&Sparse K &28.5(13.2)&63.4(28.5) &163.6 (102.9) &218.6 (129.9) \\
&IF-PCA &50.8(5.2) &75.4(16.5)&126.9(61.2)&209.3(130.5)\\
\hline
\multirow{4}{*}{1.0}
&SAS\_gs &10.5 (3.7)  &10.2 (3.4)&12.4 (3.7) &22.7(19.6)  \\
&SAS\_gss &11.7(3.9)&12.5 (4.1) & 17.1 (12.8)&100.2 (84.8) \\
&Sparse K &13.6 (10.8)&49.7 (38.2) &204.88 (104.5) &265 (165.1) \\
&IF-PCA &49.3(4.0) &67.9(14.1) &124.8(53.3) &226.3(146.7)\\
\hline
\end{tabular}

\caption{Comparison of SAS Clustering with Sparse K-means  and IF-PCA in the setting of \secref{sim_same_sigma}.  Reported is the averaged symmetric difference over 50 repeats, with the standard deviation in parentheses.}
\label{tab:sim_same_sigma_b}
\end{table}
\addtocounter{table}{-1}

\subsection{A more difficult situation (different covariances)} 
\label{sec:sim_diff_sigma}

In both \secref{sim_identity} and \secref{sim_same_sigma}, the three groups have the same covariance matrix.  In this section, we continue comparing our approach with Sparse K-means and IF-PCA-HCT under an even more difficult situation, where the mean vectors are the same as in \secref{sim_same_sigma} with $\delta_\mu=1.0$, but now the covariances are different: $\bSigma_1$, $\bSigma_2$ and $\bSigma_3$ are random matrices with eigenvalues in $[1,2]$, $[2,3]$ and $[3,4]$, respectively.  
We used 50 repeats in this simulation.  The results, reported in \tabref{sim_diff_sigma}, are consistent with the results of \secref{sim_same_sigma}: our method clearly outperforms Sparse K-means and IF-PCA-HCT, both in terms of clustering and feature selection. 

\begin{table}[h]
\centering\footnotesize
\begin{tabular}{ccccc}
\hline
Method & SAS\_gs & SAS\_gss & Sparse K-means & IF-PCA\\
\hline 
Rand index & 0.920 (0.054) & 0.858 (0.098) &0.710 (0.022)  & 0.668 (0.041) \\
\hline
$|S_* \symd \hat S|$  &8.7 (3.8)  &13.0 (7.9)  &297.2 (75.6)  & 118.6 (56.8) \\
\hline
\end{tabular}
\caption{Comparison of SAS Clustering with Sparse K-means  and IF-PCA in the setting of \secref{sim_diff_sigma}.  Reported are the Rand index and symmetric difference, averaged over 50 repeats.  The standard deviations are in parentheses.}
\label{tab:sim_diff_sigma}
\end{table}

\begin{figure}[h]
\centering\footnotesize
\subfigure[True clustering]{
	\includegraphics  [scale=0.35]{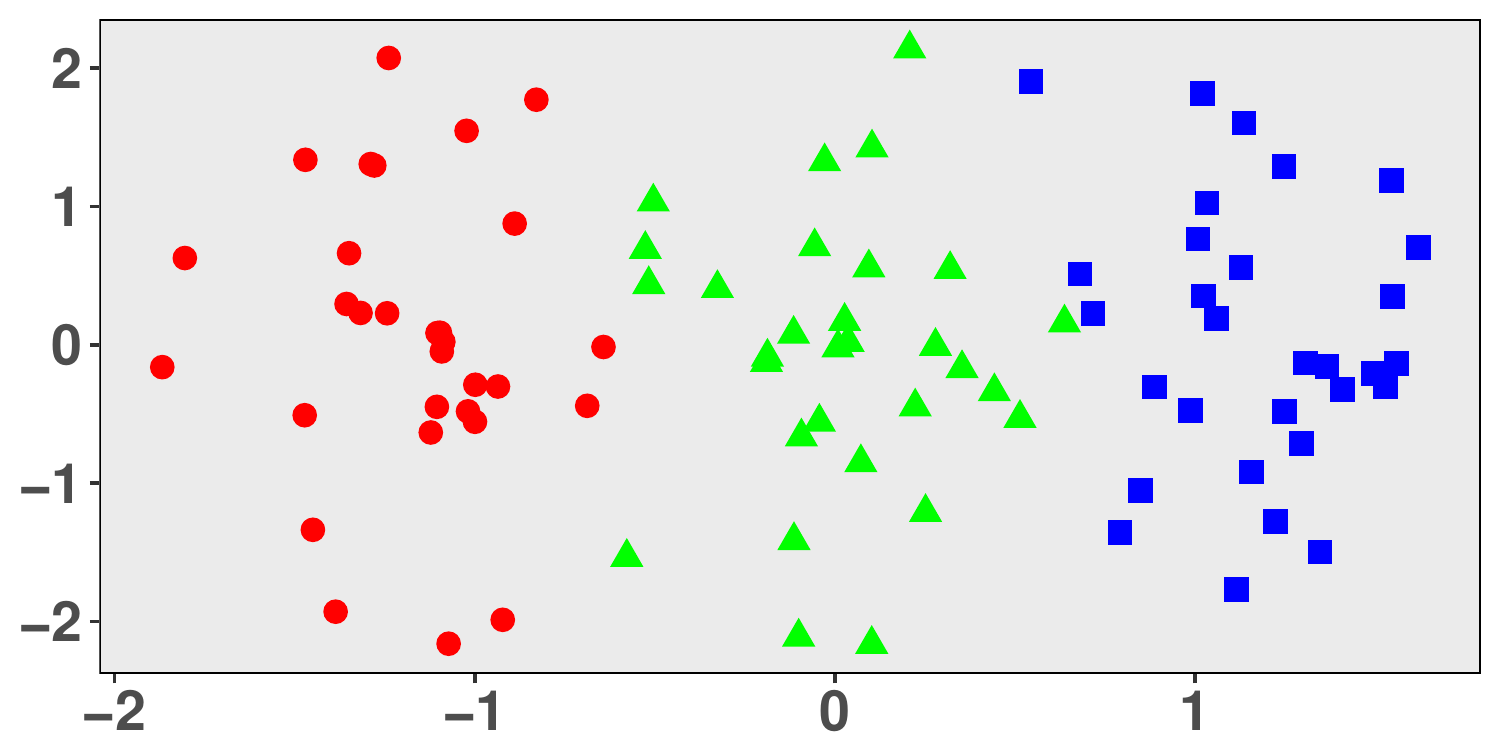}
	\label{fig:spca}}\hspace{1em}
\subfigure[True clustering$^*$]{
	\includegraphics  [scale=0.35]{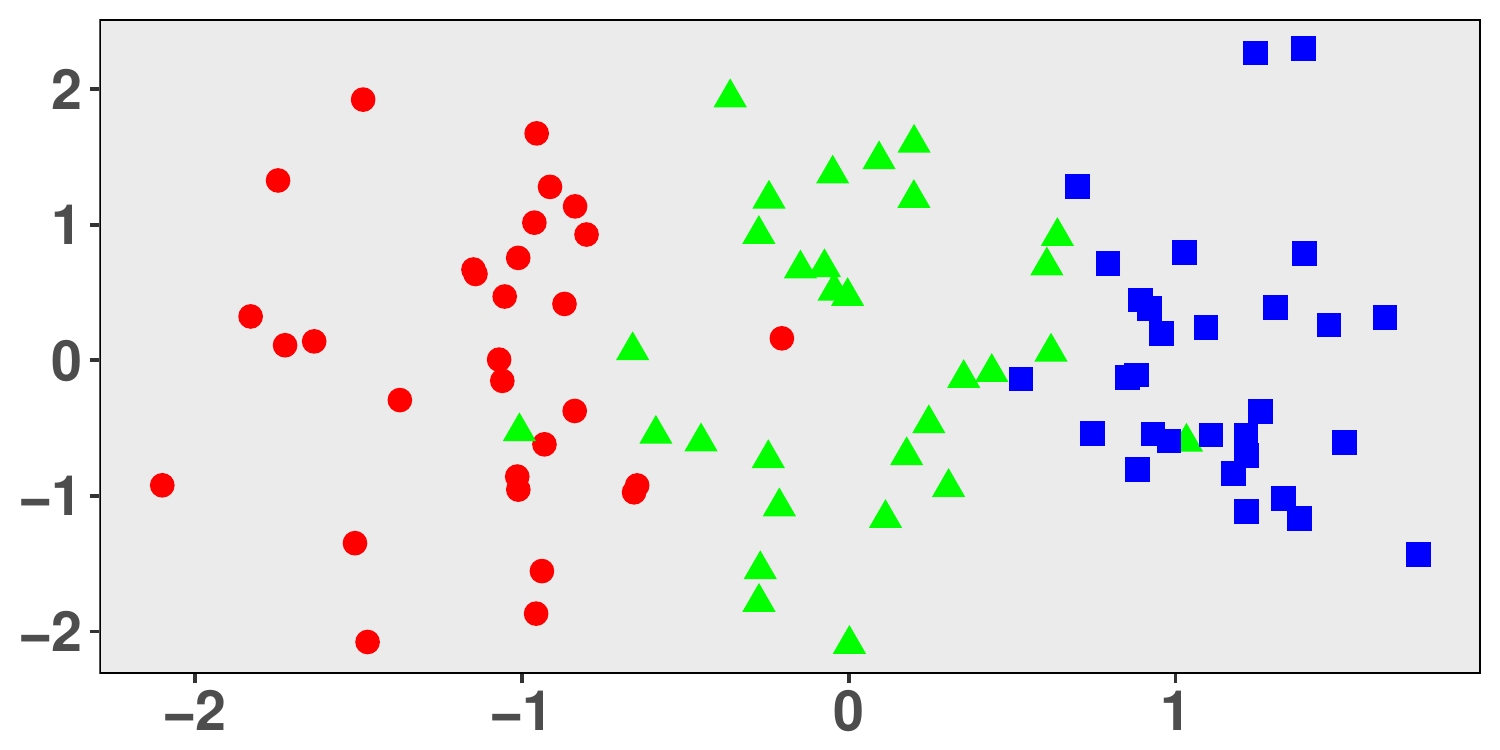}
	\label{fig:pca}}
	
\subfigure[Clustering by SAS\_gs]{
	\includegraphics  [scale=0.35]{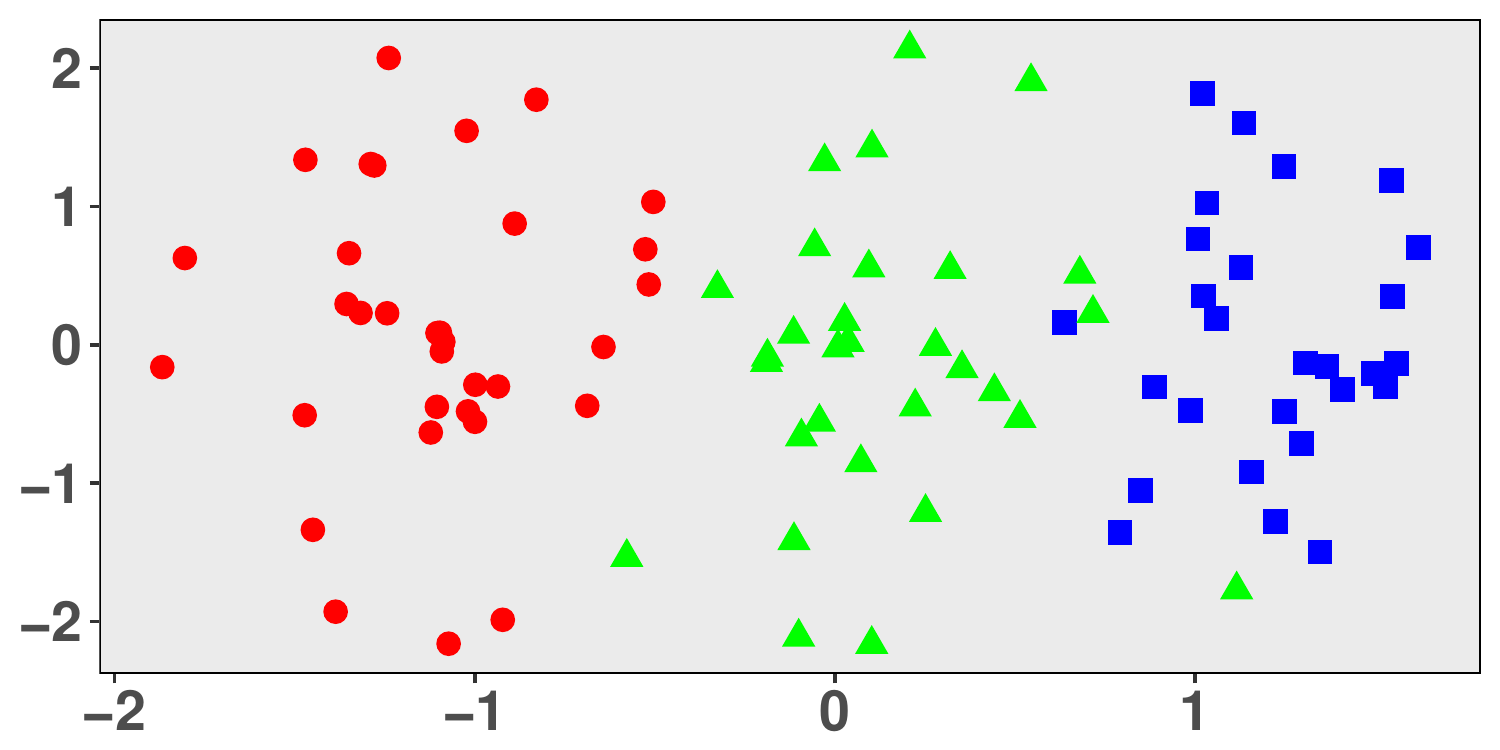}
	\label{fig:SAS_gs}}\hspace{1em}	
\subfigure[Clustering by SAS\_gss]{
	\includegraphics  [scale=0.35]{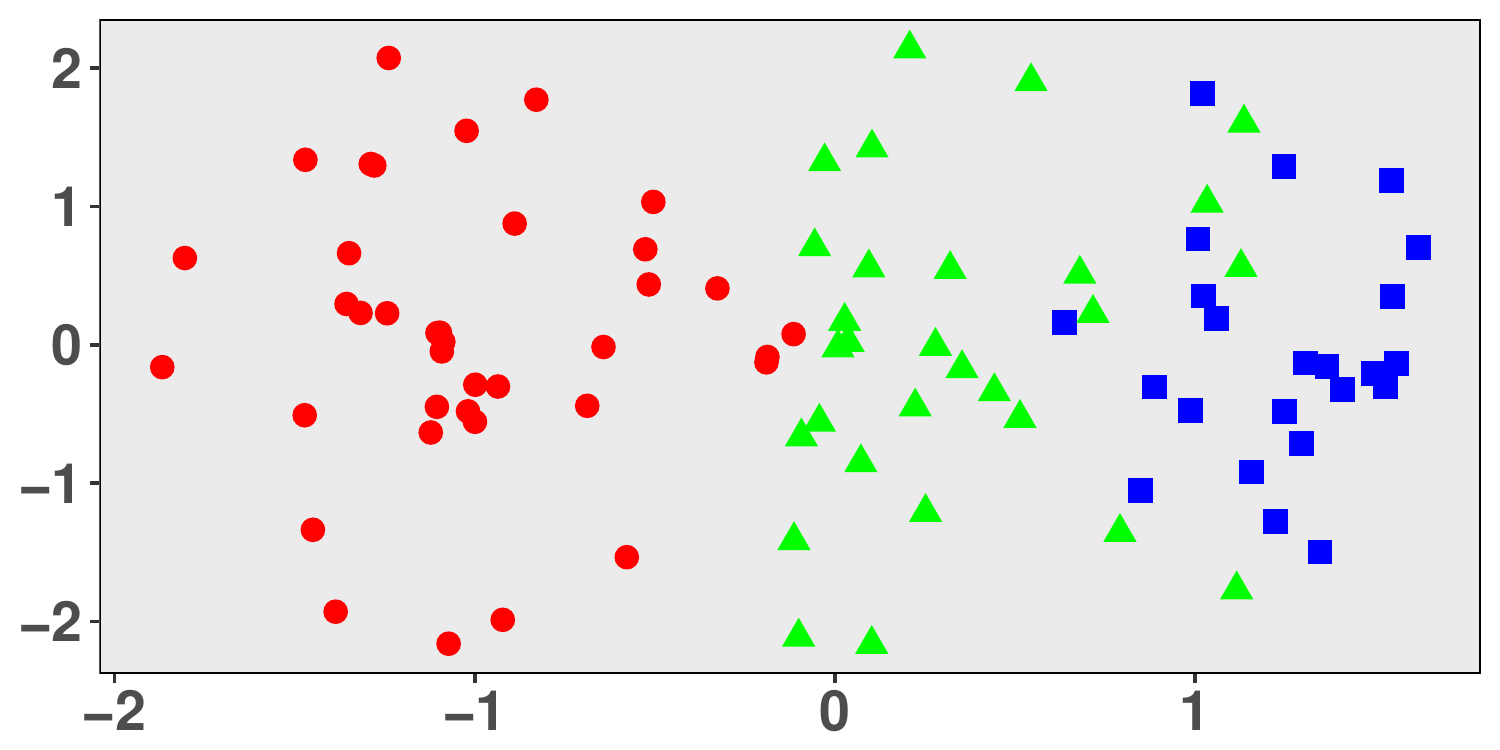}
	\label{fig:SAS_gss}}

\subfigure[Clustering by Sparse K-means]{
	\includegraphics  [scale=0.35]{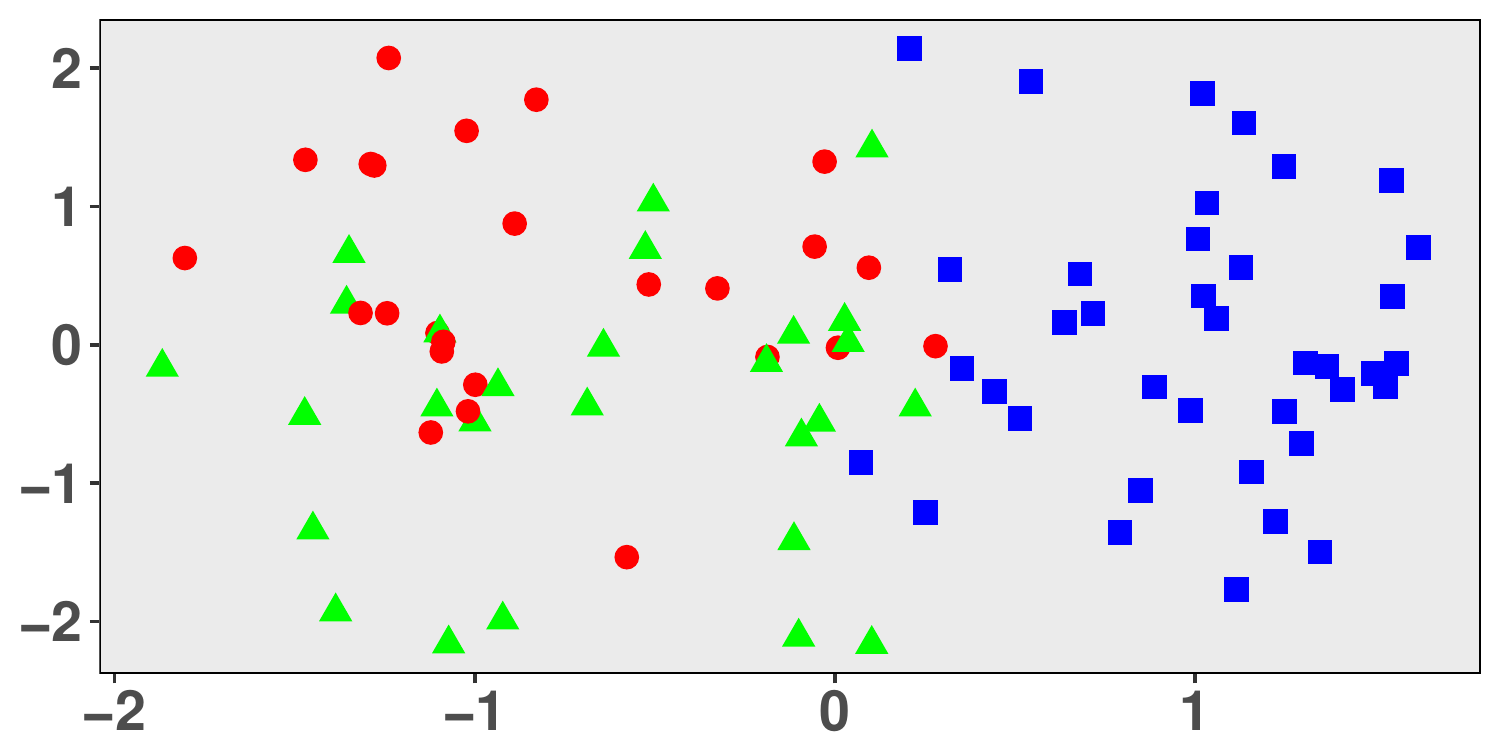}
	\label{fig:Skmeans}}\hspace{1em}
\subfigure[Clustering by IF-PCA]{
	\includegraphics  [scale=0.35]{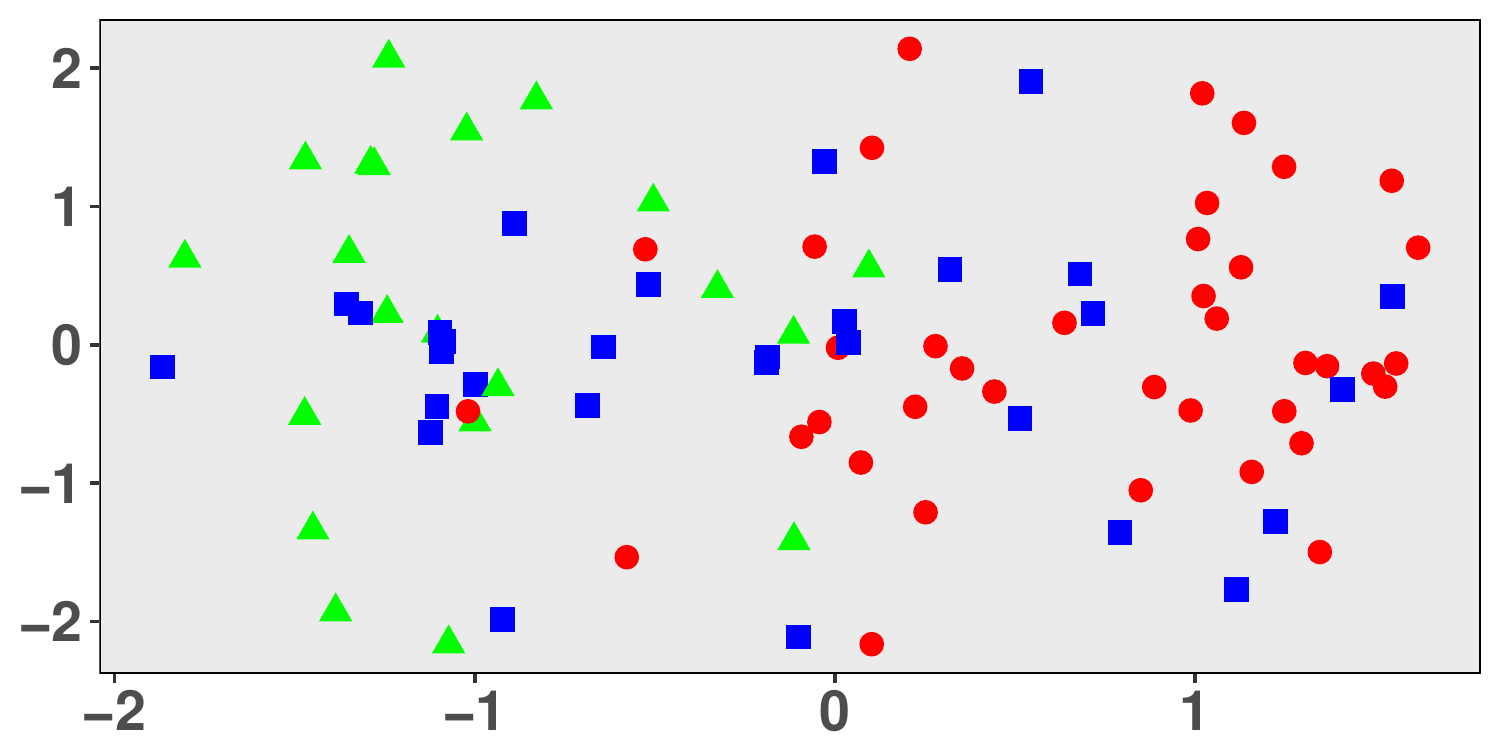}
	\label{fig:IF-PCA}}
\caption{Projection of a dataset from \secref{sim_diff_sigma} onto the first two principal components of the data submatrix, where only the first 50 columns are kept.
$^*${\footnotesize Different from the other 5 subfigures, here the data points are projected onto the first two principal components of the whole data matrix.}
}
\label{fig:pca-visual}
\end{figure}

Notice that the 3 clusters are well separated in the first 50 features as can be seen from the construction of the data, but when 450 noise features are present in the datasets, the task of clustering becomes difficult. See \figref{pca} as an example where we project a representative dataset onto the first two principal components of the whole data matrix. However, if we are able to successfully select out the first $50$ features and apply classical clustering algorithms, then we are able to achieve better results. See \figref{spca}, where we project the same dataset onto the first two principal components of the data submatrix consisting of the first 50 columns (features). 
To illustrate the comparisons, we also plot in \figref{pca-visual} the clustering results by these three methods.

\subsection{Clustering non-euclidean data}
\label{sec:hamming}
In the previous simulations, all the datasets were Euclidean. 
In this section, we apply our algorithm on categorical data (with Hamming distance) and compare its performance with Sparse K-medoids\footnote{We modified the function of Sparse K-means in the R package `sparcl', essentially replacing K-means with K-medoids, so that it can be used to cluster categorical data.}. 
In this example, we generate $3$ clusters with $30$ data points each from three different distributions on the Hamming space of dimension $p$.  
Each distribution is the tensor product of Bernoulli distributions with success probabilities $q_a \in [0,1]$ for $a \in [p]$.
For the first distribution, $q_a = q$ for $1 \le a \le 5$ and $q_a = 0.1$ otherwise.  
For the second distribution, $q_a = q$ for $6 \le a \le 10$ and $q_a = 0.1$ otherwise.  
For the third distribution, $q_a = q$ for $11 \le a \le 15$ and $q_a = 0.1$ otherwise.  
See \tabref{categorical}, where we compare these two methods in terms of Rand index for various combination of $q$ and $p$. 
Each situation was replicated 50 times. As can be seen from the table, SAS Clustering significantly outperforms Sparse K-medoids in most situations. 
We examined why, and it turns out that Sparse K-medoids works well if the tuning parameter $s$ in equation \eqref{eq:WT's condition} is given, but it happens that the gap statistic often fails to give a good estimate of $s$ in this categorical setting.  We are not sure why.

\begin{table}[h]
\centering\footnotesize
\begin{tabular}{lccccc}
\hline
$q$ & methods & p = 30 & p = 60& p = 100& p = 200\\
\hline
\multirow{2}{*}{0.6} 
& SAS\_gs & 0.878 (0.060)   & 0.872 (0.042) & 0.864 (0.057) &  0.863 (0.053)  \\
& Sparse K-medoids &0.694 (0.045) &0.663 (0.054)&0.654 (0.049) &0.639 (0.044)\\
\hline
\multirow{2}{*}{0.7}
& SAS\_gs & 0.954 (0.023)   & 0.960 (0.026) & 0.942 (0.026) &  0.948 (0.033)  \\
& Sparse K-medoids &0.807 (0.126) &0.763 (0.077)&0.716 (0.060) &0.686 (0.062)\\
\hline
\multirow{2}{*}{0.8}
& SAS\_gs & 0.989 (0.011)   & 0.984 (0.019) & 0.983 (0.019) &  0.978 (0.021)  \\
& Sparse K-medoids &0.946 (0.090) &0.889 (0.099)&0.846 (0.100) &0.787 (0.093)\\
\hline
\multirow{2}{*}{0.9}
& SAS\_gs &0.998 (0.005)   &0.999 (0.003) & 0.997 (0.007) &  0.997 (0.006)  \\
& Sparse K-medoids &0.997 (0.006) &0.994 (0.036)&0.983 (0.044) &0.966 (0.065)\\
\hline
\end{tabular}
\caption{Comparison results for \secref{hamming}. The reported values are the mean (and standard error) of the Rand indexes over 50 simulations. 
}
\label{tab:categorical}
\end{table}

\subsection{Comparisons as the number of clusters $\kappa$ increases}
\label{sec:diff_kappa}
\begin{figure}[h]
\centering\footnotesize
\subfigure[30 observations in each cluster]{
	\includegraphics  [scale=0.25]{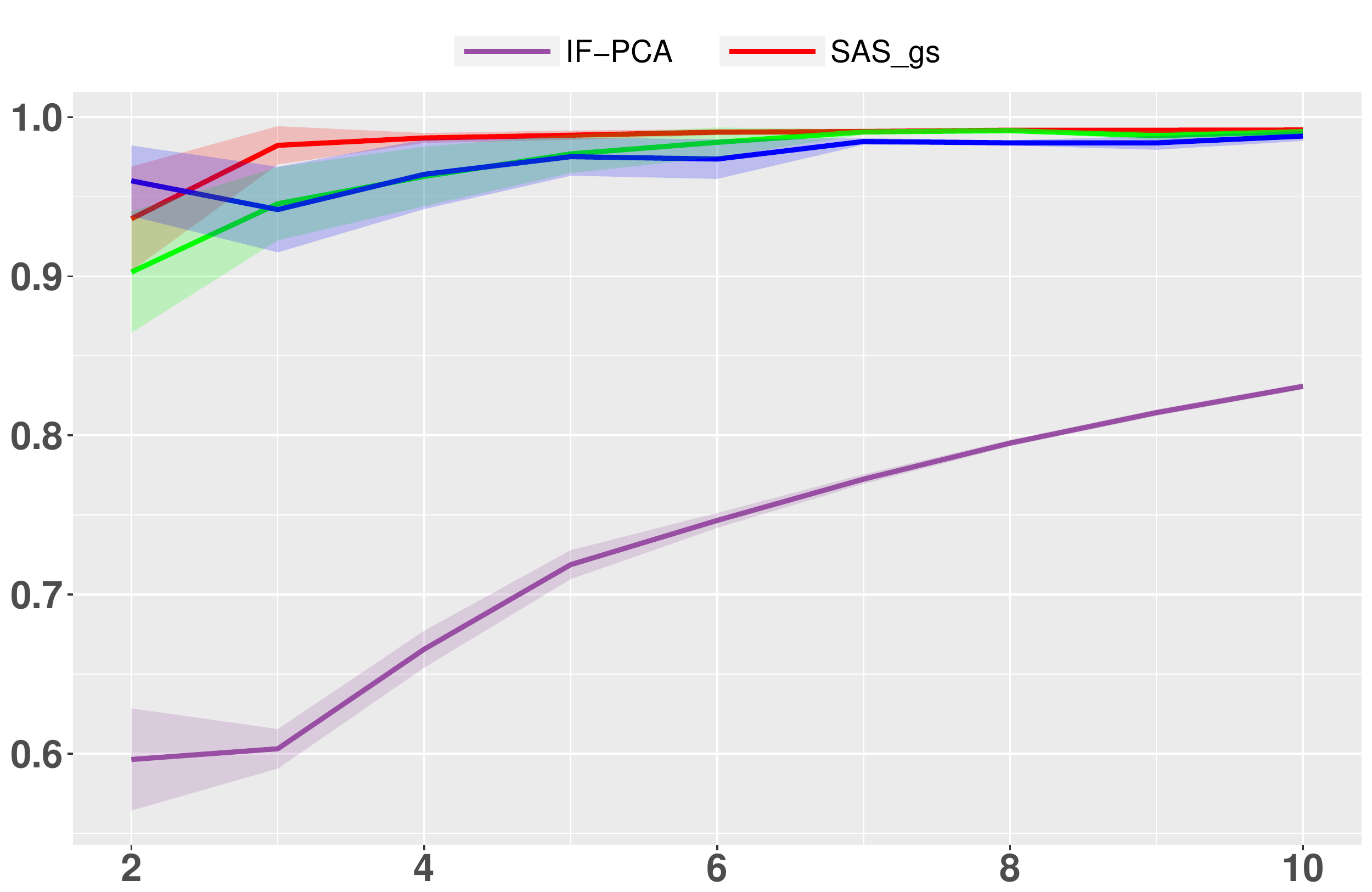}\hspace{1em}
	\includegraphics  [scale=0.25]{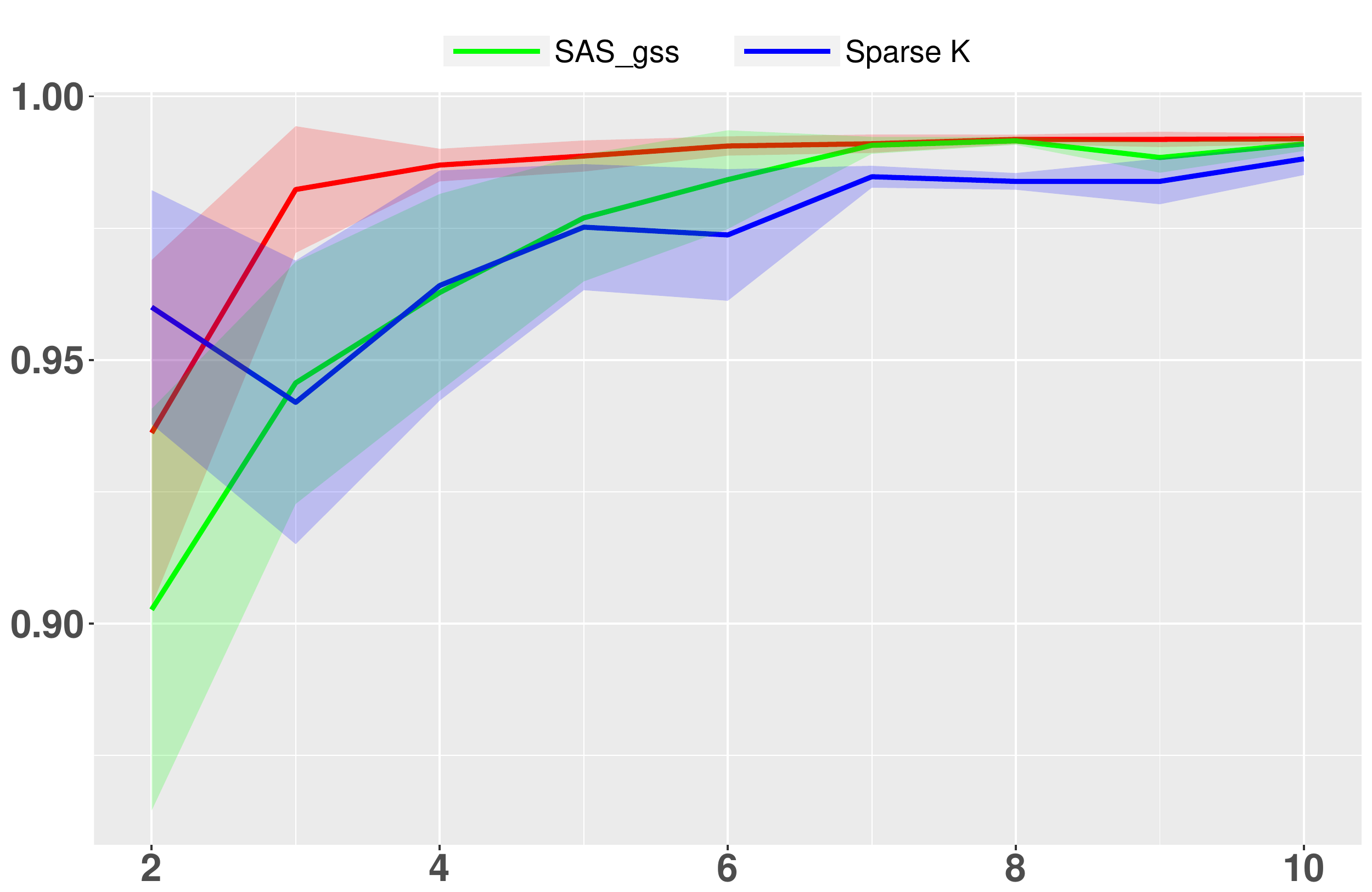}
	\label{fig:kappa_a}}
\subfigure[20 observations in each cluster]{
	\includegraphics  [scale=0.25]{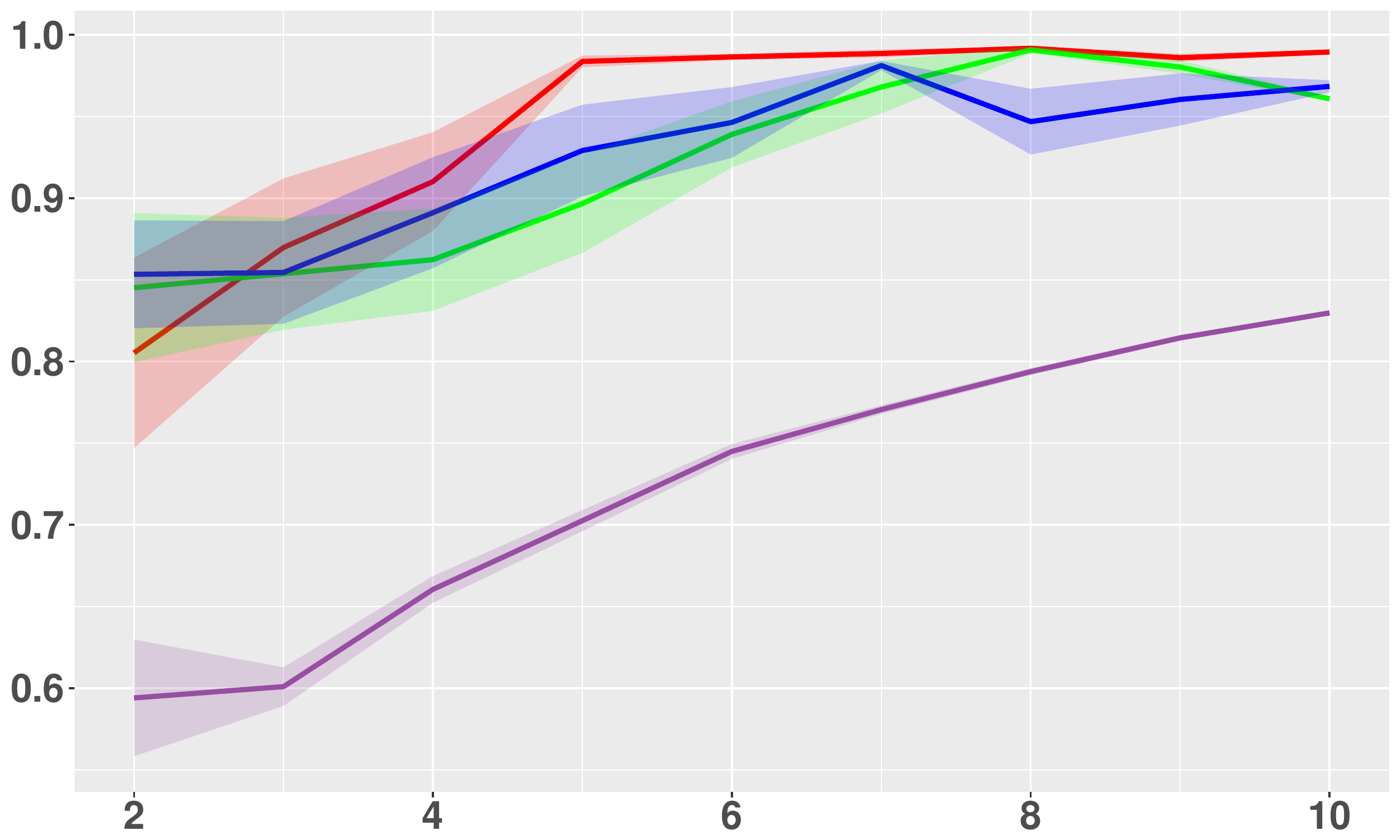}\hspace{1em}
	\includegraphics  [scale=0.25]{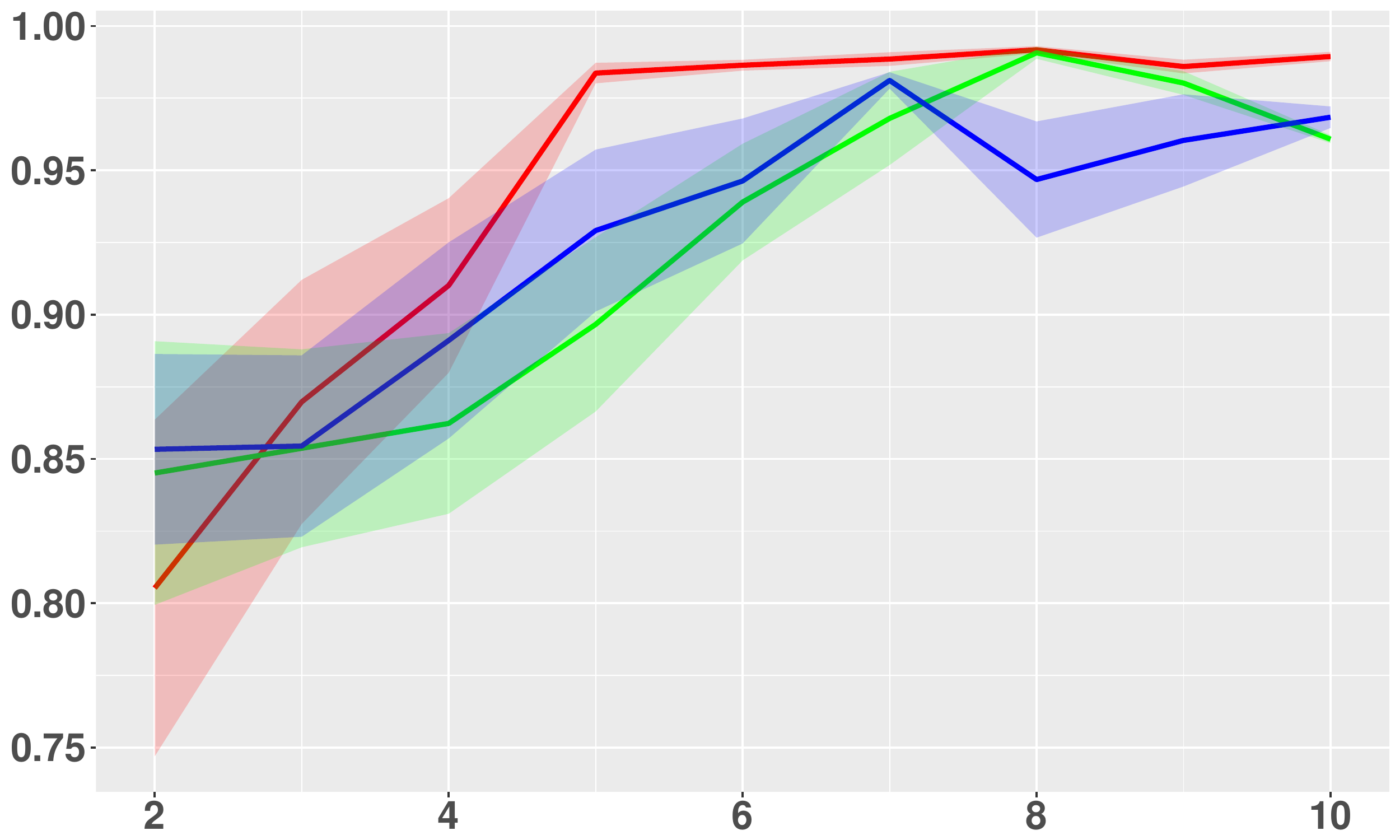}
	\label{fig:kappa_b}}
	
\caption{Comparison of SAS Clustering with Sparse K-means and IF-PCA in the setting of \secref{diff_kappa}. Reported are the means (and confidence intervals) of Rand indexes ($y$-axis) as the number of clusters, $\kappa$ ($x$-axis), increases. For each sub-figure, we separately put the same plot on the right with the results of SAS Clustering and Sparse K-means only, which clearly outperform IF-PCA.}
\label{fig:diff_kappa}
\end{figure}

In Sections \ref{sec:sim_identity} -- \ref{sec:hamming}, we have fixed the number of clusters to be $3$ and considered the effects of cluster separation $(\mu, q)$, sparsity ($p$) and cluster shape (Identity covariance, same and different covariance matrices across groups) in the comparisons. In this section, we continue to compare our approach with Sparse K-means and IF-PCA-HCT as the number of clusters $\kappa$ increases from $2$ to $10$. The set-up here is different from the above sections. We sample $\kappa$ sub-centers from a 50-variate normal distribution $\cN({\bf 0},0.4\times\bI_{50})$\footnote{The constant 0.4 was chosen to make the task of clustering neither too easy nor too difficult.} and concatenate each of the sub-centers with 450 zeros to have $\kappa$ random centers, $\bmu_1, \cdots, \bmu_\kappa$, of length $500$, which carry at least $450$ noise features. Once the centers are generated, we construct $\kappa$ clusters with $30$ ($20$ in the second set-up) observations each, sampled from respective distributions $\cN({\bmu_i}, \bI_{500})$ with $i = 1,2,\cdots, \kappa$. Each setting is repeated $50$ times. The means and confidence intervals with different $\kappa$'s are shown in \figref{kappa_a} and  \figref{kappa_b}. Once again, the results were consistent with earlier results in that SAS Clustering outperforms IF-PCA-HTC and performs at least as well as Sparse K-means with different $\kappa$'s.  We also notice that the clustering results given by all these three methods become better as $\kappa$ increases. This can be explained by the increased effective sample sizes ($30\times\kappa$ or $20\times\kappa$) as $\kappa$ increases. 

\subsection{Applications to gene microarray data}
We compare our approach with others on real data from genetics.  
Specifically, we consider the same microarray datasets (listed in \tabref{10data}) used by \cite{jin2014important} to evaluate their IF-PCA method. Each of these 10 data sets consists of measurements of expression levels of $p$ genes in $n$ patients from $\kappa$ different classes (e.g., normal, diseased).  We notice from \tabref{10data} that $p$ is much greater than $n$, illustrating a high-dimensional setting. We also mention that, although the true labels are given by the groups the individuals belong to, they are only used as the \emph{ground truth} when we report the classification errors of the different methods in \tabref{microarray}. For detailed descriptions and the access to these 10 datasets, we refer the reader to \citep{jin2014important}.

In \tabref{microarray}, we report the classification errors of 10 different methods on these datasets. Among these 10 methods, the results from K-means, K-means++ \citep{arthur2007k}, hierarchical clustering, SpectralGem \citep{lee2010spectral} and IF-PCA-HCT \citep{jin2014important} are taken from  \citep{jin2014important}. We briefly mention that K-means++ is Lloyd's algorithm for K-means but with a more careful initialization than purely random; hierarchical clustering is applied to the normalized data matrix $X$ directly without feature selection; and SpectralGem is PCA-type method.  In addition to these 5 methods, we also include 3 other methods: AHP-GMM \citep{wang2008variable}, which is an adaptively hierarchically penalized Gaussian-mixture-model based clustering method, Regularized K-means \citep{sun2012regularized}, and Sparse K-means \citep{witten2010framework}.

We can offer several comments. First, our method is overall comparable to Sparse K-means and IF-PCA, which in general outperform the other methods.  It is interesting to note that SAS\_gss outperforms SAS\_gs on a couple of datasets.  However, we caution the reader against drawing hard conclusions based on these numbers, as some of the datasets are quite small.  For example, the Brain dataset has $\kappa = 5$ groups and a total sample size of $n=42$, and is very high-dimensional with $p = 5,\!597$.
Second, for Breast Cancer, Prostate Cancer, SRBCT and SuCancer, all methods perform poorly with the best error rate exceeding $31\%$. However, we note that even when the task is classification where class labels in the training sets are given, these data sets are still hard for some well-known classification algorithms \citep{dettling2004bagboosting, yousefi2010reporting}. 
%
Third, we notice that in \citep{sun2012regularized}, clustering results of the Leukemia and Lymphoma datasets have also been compared. The error rate on Lymphoma given by Regularized K-means in \citep{sun2012regularized} is the same as reported here, however, the error rate on Leukemia is smaller than the result reported here. This is due to the fact that they applied preprocessing techniques to screen out some inappropriate features and also imputed the missing values using 5 nearest neighbors on this data set. Interestingly, \cite{wang2008variable} also reported a better error rate on SRBCT data using their AHP-GMM method. However, they split the data into training set and testing set, fit the penalized Gaussian mixture model and report the training error and testing error respectively.

\begin{table}[h]
\centering\footnotesize
\begin{tabular}{ccccc}
\hline
\# & Data Name &  $\kappa$ & $p$ & $n$ (with sample size from each cluster)\\
\hline
 1 & Brain & 5 &  5597 & 42 (10+10+10+4+8)\\
 2 & Breast & 2&22215 & 276 (183+93)\\
 3 &Colon &  2 &  2000& 62 (22+40)\\
 4& Lung &2& 12533 &181 (150+31)\\
 5&Lung(2)&2&12600 & 203 (139+64)\\
 6&Leukemia & 2& 3571& 72 (47+25)\\
 7&Lymphoma &3&4026&62 (42+9+11)\\
 8& Prostate&2&6033&102 (50+52)\\
 9&SRBCT&4&2308&63 (23+8+12+20)\\
 10&SuCancer & 2 &7909&174 (83+91)\\
 \hline
\end{tabular}
\caption{10 gene microarray datasets.} 
\label{tab:10data}
\end{table}

\begin{sidewaystable}
    \centering
\centering\footnotesize
\tabcolsep=0.08cm
\begin{tabular}{ccccccccccc}
\hline
Data set & K-means & K-means++ &Hier &SpecGem &IF-PCA& AHP-GMM & RKmeans & Sparse K& $\rm SAS_{gs}$& $\rm SAS_{gss}$\\
\hline 
 Brain  & .286 &.472 &.524 & \bf.143 & .262 & .214  & .262         & .190& .310&.310 \\
 Breast &.442&.430&.500&\bf.438 &.406 & .460  & .442 &.449&.485&.445  \\
 Colon &  .443 & .460 &  .387 &.484 & .403& \bf  .129  & .355   & .306 &\bf  .129&.403\\
Lung &.116&.196&.177&.122&\bf.033 & .116  & .094  &.122&.099&.099  \\
Lung(2)&.436&.439&.301&.434&\bf.217 & .438  & \bf.217  &.315&.315&.315  \\
Leukemia & .278 &.257&.278 &.292  &  .069 & \bf.028  & .347 &\bf.028&\bf .028&\bf .028   \\
Lymphoma& .387&.317&. 468& .226 & .065& .484  & \bf.016  & \bf.016&\bf.016&\bf.016  \\
 Prostate&.422&.432&.480&.422 &.382& .422  & .441  & \bf.373&.431&.431   \\
SRBCT&.556&.524&.540&.508 &.444& .476  & .556  &\bf.317&  .460&.365\\
 SuCancer & .477 &  .459 & .448 &.489 &\bf.333  & .477  & .477  & .477 &.483&.483 \\
\hline
\end{tabular}
\caption{Comparison of SAS Clustering with other clustering methods on 10 gene microarray datasets.  (In {\bf bold} is the best performance.)} 
\label{tab:microarray}
\end{sidewaystable}

\section{Conclusion}
\label{sec:conclusion}

We presented here a simple method for feature selection in the context of sparse clustering.  The method is arguably more natural and simpler to implement than COSA or Sparse K-means.  At the same time, it performs comparably or better than these methods, both on simulated and on real data.  

At the moment, our method does not come with any guarantees, other than that of achieving a local minimum if the iteration is stopped when no improvement is possible.  Just like other iterative methods based on alternating optimization, such as Lloyd's algorithm for K-means, proving a convergence to a good local optimum (perhaps even a global optimum) seems beyond reach at the moment.
COSA and Sparse K-means present similar challenges and have not been analyzed theoretically.  IF-PCA has some theoretical guarantees developed in the context of a Gaussian mixture model \citep{jin2014important} --- see also \cite{jin2015phase}.  More theory for sparse clustering is developed in \citep{azizyan2013,verzelen2014detection,chan2010using}.




\bibliographystyle{elsarticle-num}
\bibliography{hill-climbing}

\begin{thebibliography}{10}
\expandafter\ifx\csname url\endcsname\relax
  \def\url#1{\texttt{#1}}\fi
\expandafter\ifx\csname urlprefix\endcsname\relax\def\urlprefix{URL }\fi
\expandafter\ifx\csname href\endcsname\relax
  \def\href#1#2{#2} \def\path#1{#1}\fi

\bibitem{ESL}
T.~Hastie, R.~Tibshirani, J.~Friedman, {The elements of statistical learning},
  Springer, New York, 2009.

\bibitem{friedman2004clustering}
J.~H. Friedman, J.~J. Meulman, Clustering objects on subsets of attributes
  (with discussion), Journal of the Royal Statistical Society: Series B
  (Statistical Methodology) 66~(4) (2004) 815--849.

\bibitem{witten2010framework}
D.~M. Witten, R.~Tibshirani, A framework for feature selection in clustering,
  Journal of the American Statistical Association 105~(490).

\bibitem{tibshirani2001estimating}
R.~Tibshirani, G.~Walther, T.~Hastie, Estimating the number of clusters in a
  data set via the gap statistic, Journal of the Royal Statistical Society:
  Series B (Statistical Methodology) 63~(2) (2001) 411--423.

\bibitem{sun2012regularized}
W.~Sun, J.~Wang, Y.~Fang, et~al., Regularized k-means clustering of
  high-dimensional data and its asymptotic consistency, Electronic Journal of
  Statistics 6 (2012) 148--167.

\bibitem{ghosh2002mixture}
D.~Ghosh, A.~M. Chinnaiyan, Mixture modelling of gene expression data from
  microarray experiments, Bioinformatics 18~(2) (2002) 275--286.

\bibitem{liu2002}
J.~S. Liu, et~al., Bayesian clustering with variable and transformation
  selections, Bayesian Statistics 7 (2003) 245--275.

\bibitem{tamayo2007metagene}
P.~Tamayo, et~al., Metagene projection for cross-platform, cross-species
  characterization of global transcriptional states, Proceedings of the
  National Academy of Sciences 104~(14) (2007) 5959--5964.

\bibitem{pan2007penalized}
W.~Pan, X.~Shen, Penalized model-based clustering with application to variable
  selection, The Journal of Machine Learning Research 8 (2007) 1145--1164.

\bibitem{wang2008variable}
S.~Wang, J.~Zhu, Variable selection for model-based high-dimensional clustering
  and its application to microarray data, Biometrics 64~(2) (2008) 440--448.

\bibitem{guo2010pairwise}
J.~Guo, E.~Levina, G.~Michailidis, J.~Zhu, Pairwise variable selection for
  high-dimensional model-based clustering, Biometrics 66~(3) (2010) 793--804.

\bibitem{xie2008penalized}
B.~Xie, et~al., Penalized model-based clustering with cluster-specific diagonal
  covariance matrices and grouped variables, Electronic journal of statistics 2
  (2008) 168.

\bibitem{fraley2006mclust}
C.~Fraley, A.~E. Raftery, Mclust version 3: an r package for normal mixture
  modeling and model-based clustering, Tech. rep., DTIC Document (2006).

\bibitem{jin2014important}
J.~Jin, W.~Wang, Important feature pca for high dimensional clustering, arXiv
  preprint arXiv:1407.5241.

\bibitem{jin2015phase}
J.~Jin, Z.~T. Ke, W.~Wang, Phase transitions for high dimensional clustering
  and related problems, arXiv preprint arXiv:1502.06952.

\bibitem{azizyan2013}
M.~Azizyan, A.~Singh, L.~Wasserman, Minimax theory for high-dimensional
  gaussian mixtures with sparse mean separation, Neural Information Processing
  Systems (NIPS).

\bibitem{verzelen2014detection}
N.~Verzelen, E.~Arias-Castro, Detection and feature selection in sparse mixture
  models, arXiv preprint arXiv:1405.1478.

\bibitem{chan2010using}
Y.-b. Chan, P.~Hall, Using evidence of mixed populations to select variables
  for clustering very high-dimensional data, Journal of the American
  Statistical Association 105~(490) (2010) 798--809.

\bibitem{kernighan1970efficient}
B.~W. Kernighan, S.~Lin, An efficient heuristic procedure for partitioning
  graphs, Bell system technical journal 49~(2) (1970) 291--307.

\bibitem{carson2001hill}
T.~Carson, R.~Impagliazzo, Hill-climbing finds random planted bisections, in:
  Proceedings of the twelfth annual ACM-SIAM symposium on Discrete algorithms,
  Society for Industrial and Applied Mathematics, 2001, pp. 903--909.

\bibitem{aggarwal1999fast}
C.~C. Aggarwal, J.~L. Wolf, P.~S. Yu, C.~Procopiuc, J.~S. Park, Fast algorithms
  for projected clustering, in: ACM SIGMoD Record, Vol.~28, ACM, 1999, pp.
  61--72.

\bibitem{rand1971objective}
W.~M. Rand, Objective criteria for the evaluation of clustering methods,
  Journal of the American Statistical association 66~(336) (1971) 846--850.

\bibitem{kou2014estimating}
J.~Kou, Estimating the number of clusters via the gud statistic, Journal of
  Computational and Graphical Statistics 23~(2) (2014) 403--417.

\bibitem{raftery2006variable}
A.~E. Raftery, N.~Dean, Variable selection for model-based clustering, Journal
  of the American Statistical Association 101~(473) (2006) 168--178.

\bibitem{arthur2007k}
D.~Arthur, S.~Vassilvitskii, k-means++: The advantages of careful seeding, in:
  Proceedings of the eighteenth annual ACM-SIAM symposium on Discrete
  algorithms, Society for Industrial and Applied Mathematics, 2007, pp.
  1027--1035.

\bibitem{lee2010spectral}
A.~B. Lee, D.~Luca, K.~Roeder, et~al., A spectral graph approach to discovering
  genetic ancestry, The Annals of Applied Statistics 4~(1) (2010) 179--202.

\bibitem{dettling2004bagboosting}
M.~Dettling, Bagboosting for tumor classification with gene expression data,
  Bioinformatics 20~(18) (2004) 3583--3593.

\bibitem{yousefi2010reporting}
M.~R. Yousefi, J.~Hua, C.~Sima, E.~R. Dougherty, Reporting bias when using real
  data sets to analyze classification performance, Bioinformatics 26~(1) (2010)
  68--76.

\end{thebibliography}

\end{document}